\documentclass[10pt,twocolumn,letterpaper]{article}

\usepackage[pagenumbers]{iccv} 

\usepackage{tcolorbox}
\usepackage{booktabs}
\usepackage{colortbl}
\usepackage{xcolor}
\usepackage{graphicx}

\newcommand{\name}{\textsc{VERIFY}\xspace} 

\makeatletter
\newcommand{\showfontsize}{font size：\f@size pt}
\makeatother

\newcommand{\qwen}{Qwen2.5~}
\newcommand{\qvq}{QVQ~}
\newcommand{\mulberry}{Mulberry~}
\newcommand{\llava}{LLaVA-CoT~}
\newcommand{\fouro}{GPT-4o~}
\newcommand{\gemini}{Gemini~}
\newcommand{\oone}{o1~}

\usepackage{enumitem}

\usepackage{soul}

\definecolor{iccvblue}{rgb}{0.21,0.49,0.74}
\usepackage[pagebackref,breaklinks,colorlinks,allcolors=iccvblue]{hyperref}

\def\paperID{11612} 
\def\confName{ICCV}
\def\confYear{2025}

\title{\name: A Benchmark of Visual Explanation and Reasoning for Investigating Multimodal Reasoning Fidelity}

\author{%
\begin{tabular}[t]{c}
Jing Bi$^{1}$ \quad Junjia Guo$^{1}$ \quad Susan Liang$^{1}$ \quad Guangyu Sun$^{2}$\quad Luchuan Song$^{1}$ \quad Yunlong Tang$^{1}$\\
Jinxi He$^{1}$ \quad Jiarui Wu$^{1}$ 
Ali Vosoughi$^{1}$ \quad Chen Chen$^{2}$ \quad Chenliang Xu$^{1}$
\end{tabular}\\[6pt]
$^{1}$University of Rochester \quad $^{2}$University of Central Florida\\
{\tt\small \{jing.bi, jguo40, sliang22, lsong11, yunlong.tang, ali.vosoughi, chenliang.xu\}@rochester.edu}\\
{\tt\small \{jhe44, jwu114\}@u.rochester.edu}\\
{\tt\small guangyu@ucf.edu, chen.chen@crcv.ucf.edu}
}

\begin{document}
\maketitle
\begin{abstract}
Visual reasoning is central to human cognition, enabling individuals to interpret and abstractly understand their environment. 
Although recent Multimodal Large Language Models (MLLMs) have demonstrated impressive performance across language and vision-language tasks, existing benchmarks primarily measure recognition-based skills and inadequately assess true visual reasoning capabilities. 
To bridge this critical gap, we introduce VERIFY, a benchmark explicitly designed to isolate and rigorously evaluate the visual reasoning capabilities of state-of-the-art MLLMs.
VERIFY compels models to reason primarily from visual information, providing minimal textual context to reduce reliance on domain-specific knowledge and linguistic biases.
Each problem is accompanied by a human-annotated reasoning path, making it \textbf{the first} to provide in-depth evaluation of model decision-making processes.
Additionally, we propose novel metrics that assess visual reasoning fidelity beyond mere accuracy, highlighting critical imbalances in current model reasoning patterns. 
Our comprehensive benchmarking of leading MLLMs uncovers significant limitations, underscoring the need for a balanced and holistic approach to both perception and reasoning.
For more teaser and testing, visit our \href{https://verify-eqh.pages.dev/}{project page}.
\vspace{-5mm}
\end{abstract}

\section{Introduction}
\label{sec:intro}

Visual reasoning is a key aspect of human intelligence, shaping how individuals acquire knowledge and interpret the world \cite{ittelson1960visual,kosslyn1996image, pylyshyn2003seeing}. 
As a key mechanism for concept formation and understanding, it has been the focus of extensive research~\cite{malkinski2025deep}. 
However, classic models that overlook reasoning abilities, often trained on small-scale datasets with limited access to formalized reasoning, struggle to generalize beyond specific tasks and domains \cite{lake2017building, marcus2018deep, szegedy2013intriguing,malkinski2025deep}.

In contrast, recent advances in Multimodal Large Language Models (MLLM)~\cite{li2023blip-2, zhu2023minigpt-4, bi2024eagle, liu2023llava, maaz2023video, tang2024empowering, zhang2023videollama} have achieved near-human performance in various vision-language tasks~\cite{liang2024survey,caffagni2024revolution,tang2023video,hua2024v2xum,tang2025generative}. 
Nevertheless, existing benchmarks predominantly evaluate recognition-based skills, such as object detection, image captioning, and optical character recognition (OCR), which primarily rely on perceptual abilities \cite{zang2024contextual,wu2024v,lin2024draw,dong2024benchmarkingimprovingimagecaption}.
\begin{figure}[!t]
    \centering
\includegraphics[width=0.9\linewidth]{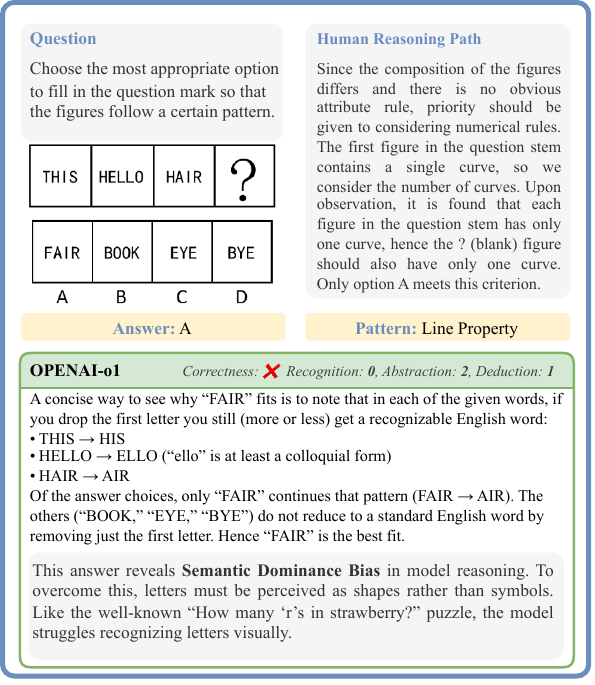}
\vspace{-2mm}
    \caption{This example demonstrates that current MLLMs primarily depend on straightforward visual signals (e.g., letters) for reasoning, frequently neglecting patterns based on other characteristics, such as shapes or line properties. \name delivers human-annotated reasoning paths to enhance the evaluation and comprehension of why and when models fail.}
        \label{fig:teaser}
        \vspace{-6mm}
\end{figure}
Recent studies have started evaluating MLLM performance in more cognitively demanding domains, including mathematical reasoning and chart-based visual understanding \cite{lu2024mathvistaevaluatingmathematicalreasoning, masry2022chartqabenchmarkquestionanswering}, which require reasoning of numerical and logical operations. 
However, these benchmarks often conflate reasoning ability with domain knowledge drawn from the underlying LLM, making it hard to isolate and accurately assess reasoning based solely on visual information.
Additionally, most evaluations emphasize accuracy as their primary metric while neglecting the decision-making quality and the depth of the reasoning process, such that even when models correctly identify the visual pattern, they often struggle to pinpoint the right answer.
This gap raises an essential research question: To what extent can MLLMs genuinely perform visual reasoning and demonstrate systematic cognitive understanding?

To address this gap, we introduce \name(\textbf{V}isual \textbf{E}xplanation and \textbf{R}easoning for \textbf{I}nvestigating
Multimodal Reasoning \textbf{F}idelit\textbf{y}), a new benchmark designed to rigorously assess the reasoning fidelity of leading MLLMs, covering both open-source and proprietary systems.
Inspired by The Raven’s Progressive Matrices\cite{barsalou1999perceptual}, \name isolates visual reasoning ability by minimizing reliance on textual input, compelling models to reason primarily through visual information.
As shown in Figure \ref{fig:teaser}, each problem includes a concise textual question and one image with answer options. This design isolates visual reasoning by requiring models to predominantly rely on non-textual cues.

By providing human-annotated reasoning paths, \name not only highlights when such biases occur but also offers insights into why models might misinterpret or overemphasize specific elements. This approach ultimately aims to foster a more balanced evaluation of visual reasoning, encouraging models to integrate both semantic and abstract visual cues for a more accurate and holistic understanding.

\begin{figure*}[!ht]
    \centering
    \includegraphics[width=1\linewidth]{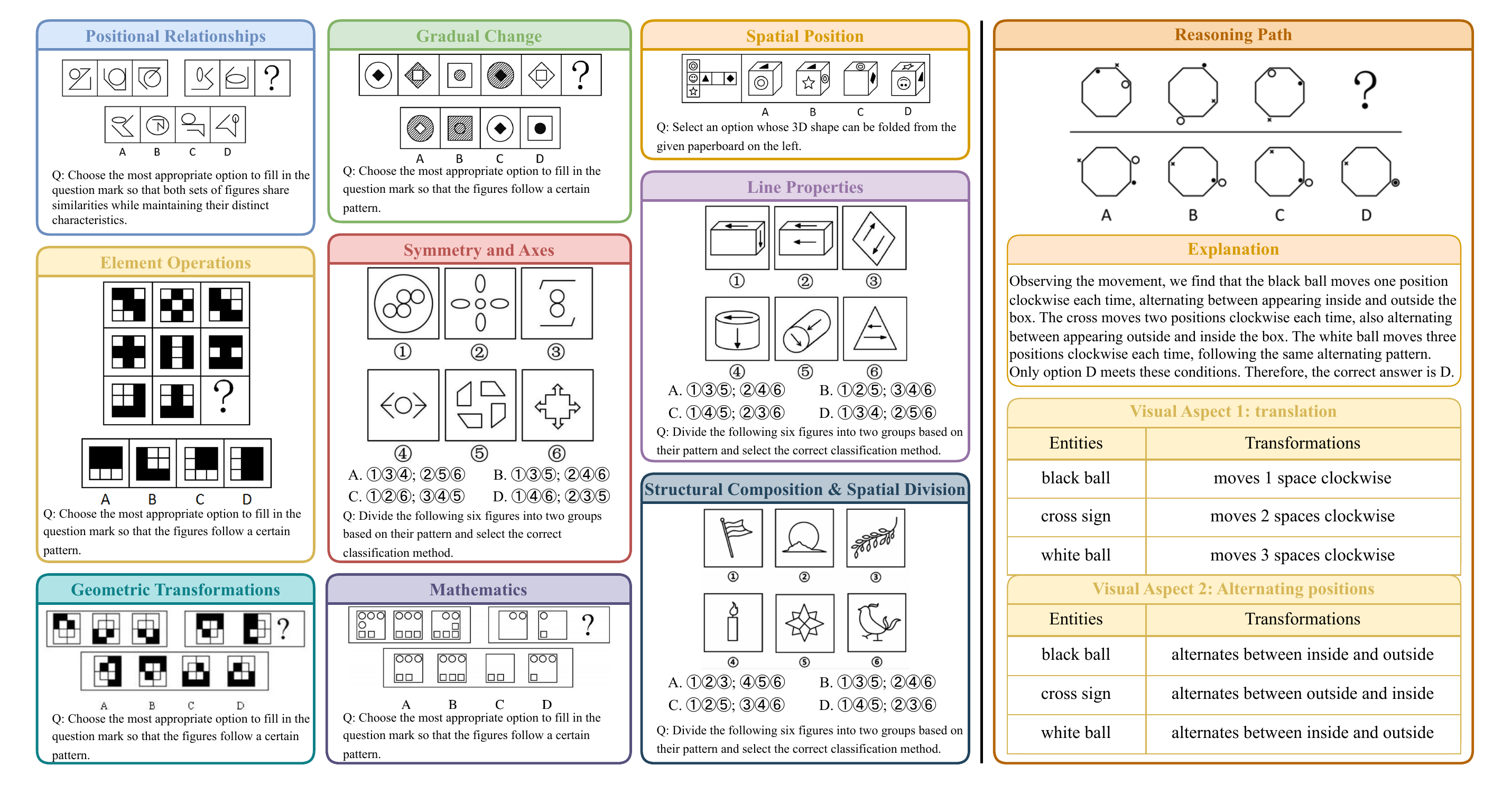}
    \caption{Categories from the VERIFY dataset cover a range of patterns, from logical operations to 3D geometry and mathematics. The right panel presents a human reasoning path, demonstrating how visual transformations, rotations, and inside-outside shifts lead to the final answer.  \textbf{We encourage readers to test these examples with MLLM models (e.g., \oone or \gemini) to assess their reasoning capabilities.}}
    \label{fig:data}
    \vspace{-5mm}
\end{figure*}

\name encompasses a diverse and challenging set of visual reasoning tasks requiring generalization and abstraction, as exemplified in Figure \ref{fig:data}, surpassing existing benchmarks in both diversity and difficulty as shown in Table \ref{tab:datasets}.

Leveraging this benchmark, we evaluate leading MLLMs from OpenAI, Google, and various most recent open-source models to provide a comprehensive analysis of their visual reasoning abilities. 
Our results reveal that even the most advanced models achieve an accuracy of only 21.7\%, which is below random chance (25\%). Furthermore, we identify shortcomings in current automated evaluation methods \cite{tyen2023llms}, noting their insufficiency in capturing nuanced differences between models.
Inspired by foundational studies in human visual reasoning \cite{barsalou1999perceptual}, we propose novel metrics designed to evaluate the reasoning process itself, providing a more refined assessment beyond accuracy. Our findings emphasize that for advanced visual reasoning models, the reasoning path holds greater importance than merely obtaining the correct answer, as models typically explore multiple reasoning paths during problem-solving.

Extensive qualitative analyses of the models’ reasoning patterns reveal a critical imbalance, where models proficient in perceiving visual details often struggle to deduce correct answers, while those excelling in abstract reasoning frequently overlook essential visual cues, highlighting a deficiency in balanced reasoning capabilities. In summary, our contributions are as follows:
\begin{itemize}[leftmargin=*,label=\tiny$\bullet$,itemsep=0pt]
\item \textbf{VERIFY}: We introduce a novel visual reasoning benchmark isolating reasoning ability by minimizing domain-specific and linguistic biases. To the best of our knowledge, it is the \emph{first} to provide human-annotated, clear, and evaluable reasoning trajectories for each question.

\item \textbf{Beyond Accuracy}: We proposal novel metrics that assess the reasoning process itself, providing a more nuanced and comprehensive evaluation beyond traditional accuracy, capturing the depth and quality of reasoning.

\item \textbf{Comprehensive Benchmarking}: \name systematically evaluates leading MLLMs on reasoning capabilities, perceptual alignment, and logical consistency, highlighting the impact of Chain-of-Thought (CoT) reasoning and the challenges in transitioning from abstraction to answer deduction, offering valuable insights.

\end{itemize}

\vspace{-3mm}

\section{Related Work}
\label{sec:related}

\noindent\textbf{Visual Reasoning.}
Visual reasoning involves comprehending and manipulating visual information to infer logical relationships \cite{agrawal2016vqavisualquestionanswering,goyal2017makingvvqamatter}. Early tasks like VQAv1 and VQAv2 \cite{agrawal2016vqavisualquestionanswering,goyal2017makingvvqamatter} focused on object-centric question answering, while CLEVR \cite{johnson2017clevr} and GQA \cite{hudson2019gqanewdatasetrealworld} introduced compositional programmatic queries for structured reasoning. Beyond static images, video-based datasets such as CoPhy and Space \cite{baradel2020cophycounterfactuallearningphysical,janny2022filteredcophyunsupervisedlearningcounterfactual,duan2021spacesimulatorphysicalinteractions} emphasize physical interactions and causal reasoning in motion.
Our work aligns with \emph{abstract} visual reasoning that does not require external domain knowledge, like RAVEN, SVRT, CVR, Bongard-HOI, and Bongard-LOGO \cite{zhang2019raven,fleuret2011comparing,zerroug2022benchmark,jiang2023bongardhoibenchmarkingfewshotvisual,nie2020bongard}, which explore relational patterns among shapes. However, as \citet{van2021much} note, many existing words have rigid configurations that hinder generalization to complex problems.\\
\vspace{-3mm}

\noindent\textbf{Reasoning Benchmarks.}
Most popular language reasoning benchmarks focus on text-based QA, mathematics, or code inference \cite{cobbe2021training,hendrycks2021measuring,srivastava2022beyond,suzgun2022challenging}, where step-by-step solutions can be programmatically verified. In the \emph{visual} domain, benchmarks like CLEVR \cite{johnson2017clevr}, GQA \cite{hudson2019gqanewdatasetrealworld}, CLEVRER \cite{yi2019clevrer}, Bongard \cite{nie2020bongard}, RAVEN \cite{zhang2019raven}, MMComposition \cite{hua2024mmcomposition}, and VidComposition \cite{tang2024vidcomposition} introduce compositional reasoning over images or videos. Recent efforts \cite{jiang2025mme} incorporate multi-modal CoT prompting, often blending external knowledge with visual cues.
Among the most relevant is MM-IQ \cite{cai2025mmiqbenchmarkinghumanlikeabstraction}, which targets abstract visual puzzles but lacks step-by-step annotations to clarify reasoning transitions. In contrast, our work provides explicit human-annotated reasoning paths, offering a more transparent, interpretable framework for fine-grained evaluation beyond the accuracy.\\

\noindent\textbf{Automatic Reasoning-Path Evaluation.}
Interest is growing in evaluating not just the accuracy a model provides, but also \emph{how} it draws its conclusions.
 \begin{enumerate*} 
 \item \textit{Embedding Methods.} \citet{golovneva2022roscoe} use embedding similarities to compare reasoning chains, effective for text but weaker for capturing geometric or visual relationships. 
 \item \textit{Symbolic Methods.} Structured parsing, such as subject-verb-object \cite{prasad2023receval} or formal proofs \cite{saparov2022language}, struggles with complex visual puzzles due to inconsistent symbolic representation of spatial relations. \item \textit{Process Reward Models.} \citet{lightman2023let} introduce PRM800K for mathematical reasoning, but training such models requires vast amounts of data, making large-scale visual adaptation costly.
 \item \textit{Prompting Methods.} LLMs can verify reasoning by prompting without ground-truth references, depending solely on internal coherence.   \cite{tyen2023llms,zeng2023mr} 
 \end{enumerate*} 
Applying these methods on visual reasoning yields poor results, as LLMs struggle to differentiate visual nuances.

\section{Data Collection}
\label{sec:data}

Our dataset stands apart from existing ones that rely on rendering engines to ensure compositionality. While synthesized data can be abundant, it often lacks the richness and complexity inherent in real-world scenarios. 
We curate our dataset primarily from publicly available questions in China’s National Civil Servants Examination, a rigorous postgraduate-level assessment. 
To mitigate any potential bias, we have incorporated questions from different provinces to ensure a diverse and comprehensive dataset.
Focusing on problems that require logical reasoning rather than simple recognition, we deliberately select complex questions demanding trial and error with multiple visual patterns. This ensures both correctness and validity while establishing a higher standard of difficulty than existing datasets. To ensure quality and clarity, we included only questions with definitive answers, filtering out ambiguous ones, those with multiple correct answers, or requiring external knowledge. Details about selection criteria are in the supplementary materials.

\subsection{Reasoning Path}

One of the key contributions of our dataset is the reasoning path, which consists of a sequence of logical steps leading to the correct answer. The common practice \cite{zhang2024multimodalchainofthoughtreasoninglanguage,cai2025mmiqbenchmarkinghumanlikeabstraction} is to rely on models to generate reasoning paths, which are then reviewed and evaluated by human annotators or an external verifier. While existing methods may work for moderately difficult tasks, they fail when faced with more challenging reasoning problems like ours. In our initial experiments, we found that state-of-the-art models struggle to generate correct reasoning paths, even as a starting point. As a result, we manually annotate each reasoning path to ensure accuracy. Each path is independently labeled by three annotators, and any discrepancies are resolved through discussion. Further details will be provided in supplementary materials.

\begin{figure*}
    \centering

    \includegraphics[width=\linewidth]{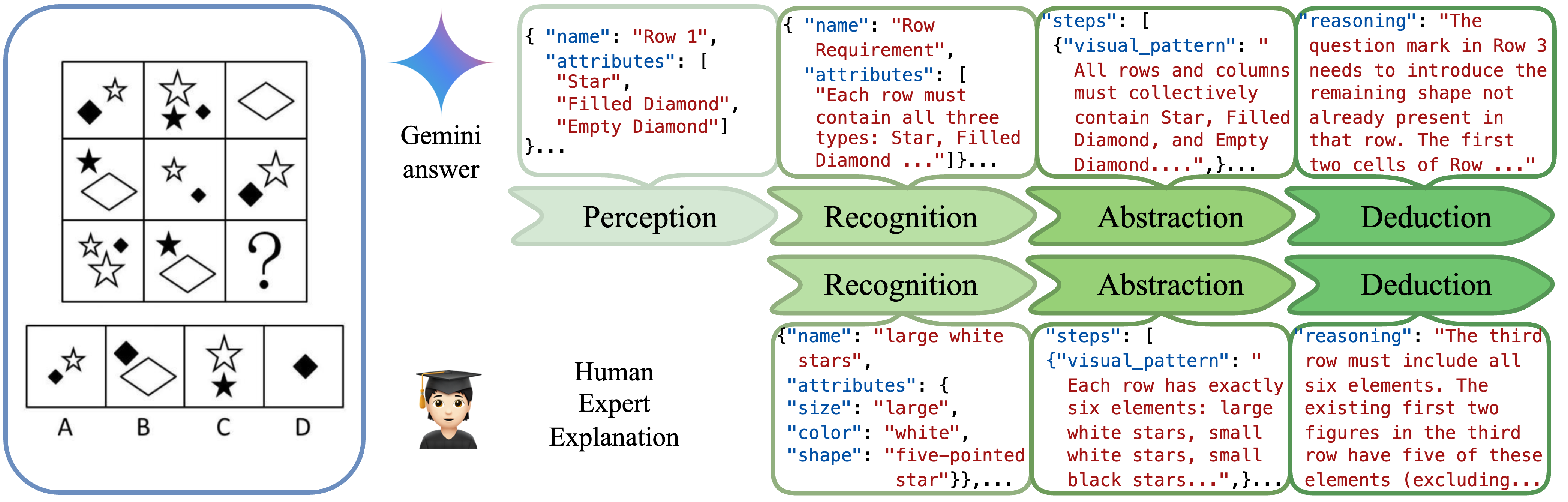}

    \caption{We divide the reasoning process into four key stages inspired by human visual reasoning: perception, recognition, abstraction, and deduction. Unlike general visual tasks, where perception involves detecting raw visual features, humans often have implicit perception because the provided visual elements are already structured for direct recognition of useful components. Even for shown complex problems, a model with strong visual abilities—like Gemini—can effectively analyze patterns and logical structures to determine the correct answer. }
    \label{fig:Reasoning Process}
    \vspace{-3mm}
\end{figure*}
\subsection{Visual Reasoning Pattern}
To solve the proposed visual reasoning problem, humans must first identify the crucial visual elements necessary for detecting patterns. However, these elements are not always immediately apparent, and we often fall prey to Semantic Dominance Bias, which causes us to focus too much on the semantic meaning of a figure rather than carefully analyzing its shape or other visual properties. To systematically approach the problem, we define the reasoning process through four stages, as shown in Figure \ref{fig:Reasoning Process}, which also enables a structured evaluation.
\begin{enumerate*}[label=\Roman*.] 
\item \emph{Perception}, where the raw visual input is processed, and features such as shapes, colors, and orientations are detected;
\item \emph{Recognition}, which involves extracting useful visual aspects from Perception, selecting the most relevant features that contribute to understanding the pattern;
\item \emph{Abstraction}, where these extracted visual aspects are used to identify patterns by filtering out unnecessary details and focusing on meaningful relationships and
\item \emph{Deduction}, the final stage, in which logical reasoning is applied to infer missing details or predict patterns based on the extracted abstractions.
\end{enumerate*}

For humans, the Perception stage is often implicit and may even be bypassed, as some initially relevant visual elements may later prove irrelevant. As a result, Recognition becomes the first explicit step, where annotators identify only the essential visual aspects and structure the reasoning path accordingly. After annotating the reasoning path, we summarize and identify key visual reasoning patterns for subsequent evaluation, as illustrated in Figure \ref{fig:data}. While MARVEL~\cite{jiang2024marvelmultidimensionalabstractionreasoning} and MM-IQ~\cite{cai2025mmiqbenchmarkinghumanlikeabstraction} provided useful insights, they proved insufficient for our dataset.

To address this limitation, we extracted refined patterns as follows that build upon existing ones while introducing unique structural distinctions. Notably, we separated symmetry from geometric transformations to enhance clarity.
\begin{enumerate*}[label=\Roman*.]
    \item \textit{Spatial Position (SP):} How elements are positioned (e.g., top-bottom, center-corner) and how they change through movement, rotation, or progression.
    \item \textit{Element Operations (EO):} Focuses on interactions between shapes, such as layering, merging, splitting, and reassembly.
    \item \textit{Geometric Transformations (GT):} Involves transformations like rotation, translation, and reflection, often occurring in sequences.
    \item \textit{Gradual Change (GC):} Tracks progressive changes in shape, size, or position, forming logical patterns.
    \item \textit{Mathematics (MA):} Observes numerical aspects, including element counts and repetitive patterns.
    \item \textit{Symmetry and Axes (SA):} Identifies different types of symmetry, including axial, central, and rotational.
    \item \textit{Line Properties (LP):} Analyzes stroke characteristics such as type, angle, continuity, and curvature.
    \item \textit{Structural Composition and Spatial Division (SC):} Examines enclosures, layering, connections, and shading to uncover deeper structural patterns.
    \item \textit{Positional Relationships (PR):} Investigates how elements relate to each other in terms of position, alignment, and directional shifts.
\end{enumerate*}

\begin{figure}[!ht]
    \centering
    \includegraphics[width=\linewidth]{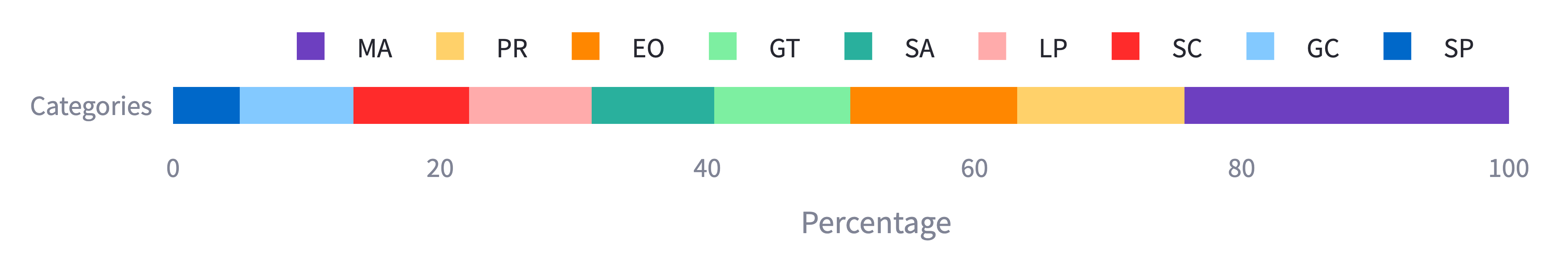}
    \caption{Category distribution of VERIFY: This chart illustrates the distribution of various mathematical and geometric concepts, with the largest segment, “Mathematics,” accounting for 24.3\%.  The remaining segments are fairly evenly distributed, reflecting a balanced emphasis on all visual patterns. }
    \label{fig:stats}
    \vspace{-2mm}
\end{figure}

\begin{table}[!h]
\centering
\resizebox{0.9\linewidth}{!}{
\begin{tabular}{l l l c l l}
\hline
 \textbf{Name} & \textbf{Size} & \textbf{Source} & \textbf{RP} & \textbf{Difficulty} & \textbf{Year} \\ \hline
SVRT~\cite{fleuret2011comparing}           & 23            & Synthesized     & \texttimes            & Easy             & 2011         \\ 
\rowcolor{gray!10}
RAVEN~\cite{zhang2019raven}          & 14,000        & Synthesized     & \texttimes            & Easy             & 2019         \\ 
ARC~\cite{chollet2019measure}            & 600           & Synthesized     & \texttimes            & Medium           & 2019         \\ 
\rowcolor{gray!10}
DOPT~\cite{webb2020learning}           & 95,200        & Synthesized     & \texttimes            & Easy             & 2020         \\ 
IQTest~\cite{lu2024mathvistaevaluatingmathematicalreasoning}         & 228           & Mixed           & \texttimes            & Medium           & 2023         \\ 
\rowcolor{gray!10}
MARVEL~\cite{jiang2024marvelmultidimensionalabstractionreasoning}         & 770           & Web             & \texttimes            & Medium           & 2024         \\ 
MM-IQ~\cite{cai2025mmiqbenchmarkinghumanlikeabstraction}          & 2,710         & Mixed           & \texttimes            & Medium           & 2025         \\ \hline
\rowcolor{gray!20}
VERIFY         & 600           & National Exam   & \checkmark            & Hard             & 2025         \\ \hline
\end{tabular}
}
\caption{Comparison of datasets from the corresponding paper. "RP" indicates the inclusion of a reasoning path, and the difficulty level is based on both the human score and the participants’ education level as reported in the original study.}
\vspace{-4mm}
\label{tab:datasets}
\end{table}

\subsection{Dataset Statistics}
The table\ref{tab:datasets} compares various datasets used for reasoning tasks, highlighting differences in size, source, difficulty, and reasoning complexity. Among them, VERIFY stands out as the most challenging due to its high quality, real-world origin (national exams), deep reasoning complexity, and rigorous difficulty level.
Compared to existing datasets, VERIFY is more challenging and diverse. While it has fewer samples than large-scale synthetic datasets, it surpasses other handcrafted datasets in both the number of samples and the diversity of reasoning patterns. Unlike synthetic datasets that focus on pattern recognition, VERIFY includes real-world, high-stakes reasoning tasks, making it a more reliable benchmark for evaluating cognitive abilities.

\section{Evaluation}
\label{sec:eval}

Given a visual reasoning problem \( q \), a generated reasoning path \( \hat{h} = \{\hat{h}_1, \dots, \hat{h}_N\} \), and a generated answer \( \hat{a} \) produced by MLLMs, our goal is to evaluate the quality of both the reasoning process and the final answer. In our case, the ground truth answer \( a \) is available, along with reference solution steps \( h = \{h_1, \dots, h_M\} \), as a reliable benchmark.

Unlike previous works that focus solely on answer matching—ignoring the entire chain of reasoning—we independently evaluate both the final answer and the reasoning path. This approach provides more informative insights into the model's reasoning process. Our empirical study reveals that models can sometimes generate the correct answer with an incorrect reasoning path or produce an incorrect answer despite following a correct reasoning process. By separately assessing these components, we gain a more comprehensive understanding of models.

\begin{figure*}[!h]
    \centering
    \includegraphics[width=0.9\textwidth]{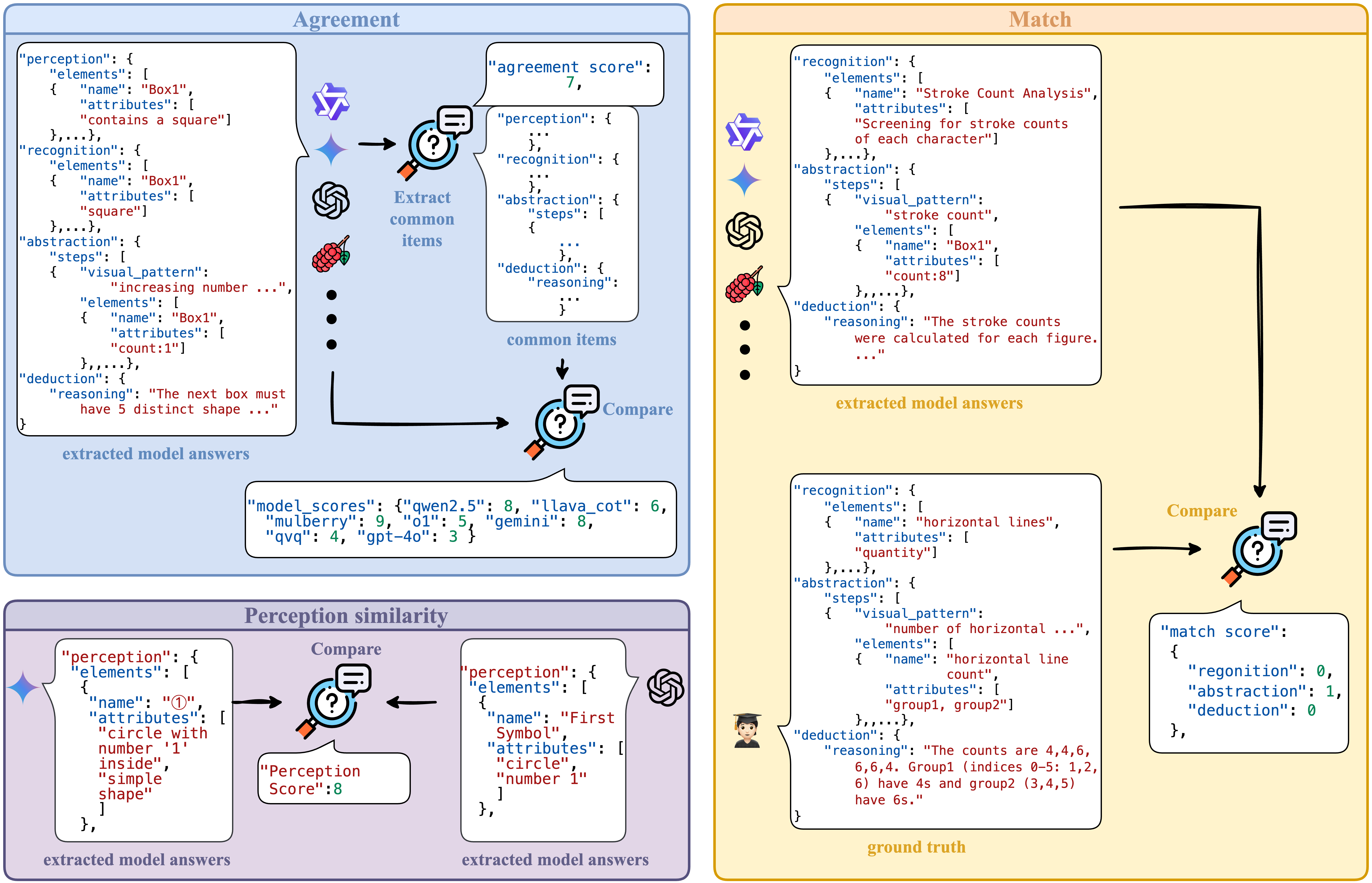}
    \caption{This diagram illustrates three proposed evaluation metrics: Agreement, which extracts common elements across model answers and compares their consistency; Match, which assesses alignment between model responses and ground truth across Recognition, Abstraction, and Deduction; and Perception Similarity, which measures the similarity of extracted perception attributes across different models.}
    \label{fig:Eval}
    \vspace{-3mm}
\end{figure*}

\subsection{Failure case of existing evaluation metrics}
\label{subsec:failure_case}
Recent studies \cite{tyen2023llms, zeng2023mr} have explored the use of open source LLMs as automated evaluators for reasoning processes, by comparing each step of the whole path against a reference reasoning trajectory. This trend aligns with the growing interest in using LLMs as judges to provide consistent and scalable evaluation frameworks.
During the early stages of designing our evaluation protocol, we tested this methodology using three of the latest high-capacity LLMs as judges: DeepSeek-R1:32B, DeepSeek-R1:70B, and LLaMA3.3-70B.
We randomly selected 40 out of 600 samples, using the o1 response as the model response, and evaluated reasoning paths by comparing them against multiple reference responses
\begin{enumerate*}[label=\Roman*.]
\item o1: The reasoning path and final answer produced by the OpenAI-o1 model.
\item Ground Truth Reasoning (GTR): A human-annotated reasoning path that correctly leads to the answer.
\item Expanded Ground Truth Reasoning (GTR+): Leverage LLM to expand and segment the GTR into three detailed stages: Recognition, Abstraction and Deduction.
\item Expanded Ground Truth Reasoning with Pattern (GTR+P): A refined version that incorporates human-labeled pattern annotations, offering additional structural insights into the reasoning process.
\end{enumerate*} Scores were assigned on a scale from 0 to 10 based on fidelity to human reasoning. The evaluation process involved assessing the alignment of model-generated reasoning paths with ground truth references. Additionally, we conducted two types of comparisons: 
\begin{enumerate*}[label=\Roman*.]
\item \textbf{r+c}: The model’s response includes both its reasoning path and the final selected choice, with the ground truth answer.
\item \textbf{r}: The model’s response includes only the reasoning path, without the final choice.
\end{enumerate*}
Based on an accuracy of 0.15 for o1 across 40 problems, several key observations emerge:
\begin{table}[h]
    \centering
    \begin{tabular}{lllcc}
        \toprule
        r1 & r2 & Model & Score (r+c) & Score \\
        \midrule
        o1 & GTR+P & deepseek-r1:32b & 7.85 & 6.40 \\
        o1 & GTR+ & deepseek-r1:32b & 7.60 & 6.80 \\
        o1 & GTR & deepseek-r1:32b & 7.80 & 6.35 \\
        GTR & GTR+P & deepseek-r1:32b & 8.45 & 7.15 \\
        GTR & GTR+ & deepseek-r1:32b & 8.70 & 8.00 \\
        o1 & GTR+P & deepseek-r1:70b & 8.10 & 7.80 \\
        o1 & GTR+ & deepseek-r1:70b & 8.35 & 8.10 \\
        o1 & GTR & deepseek-r1:70b & 7.65 & 6.60 \\
        GTR & GTR+P & deepseek-r1:70b & 8.65 & 8.65 \\
        GTR & GTR+ & deepseek-r1:70b & 8.80 & 8.65 \\
        o1 & GTR+P & llama3.3 & 8.05 & 8.20 \\
        o1 & GTR+ & llama3.3 & 8.15 & 8.35 \\
        o1 & GTR & llama3.3 & 7.95 & 8.20 \\
        GTR & GTR+P & llama3.3 & 9.05 & 9.40 \\
        GTR & GTR+ & llama3.3 & 9.25 & 9.85 \\
        \bottomrule
    \end{tabular}
    \caption{Using an LLM as a judge to compare the relevance scores of two reasoning paths often falls short in correctness. r1 and r2 are the two reasoning paths we compare against.}
    \label{tab:model_performance}
    \vspace{-5mm}
\end{table}

\noindent \textbf{Overestimation in LLM Judges.} 
Table \ref{tab:model_performance} shows a striking discrepancy: human evaluators marked only 6 out of 40 answers as correct, yet LLMs consistently assigned scores in the range of 6.35–9.85. In our experiment, we ask each LLM to compare two reasoning paths and score them on a scale from 1 to 10 based on how well they align. This disparity indicates that LLMs tend to overestimate reasoning quality, often relying on superficial similarity rather than a rigorous evaluation of logical correctness. Notably, LLaMA3.3-70B assigns comparably high scores to both GTR and its structured variants (GTR+ and GTR+P), while it awards slightly lower scores to o1 relative to the ground truth. Nonetheless, the overestimation bias persists, underscoring the challenge LLMs face in adequately penalizing flawed reasoning.

\noindent \textbf{Bias Toward Completed Choices.}
Surprisingly, LLMs tend to assign higher scores when the final choice is included (r+c), even though they should recognize that the answer is incorrect and penalize it accordingly. We suspect that including the choice makes the reasoning path feel more complete, biasing LLM judges toward a more favorable answer.

Our experiment reveals that LLMs prioritize completeness over actual correctness, favoring coherence even at the expense of reasoning accuracy. This bias results in a failure to properly penalize flawed logic, leading to overestimation and missed nuances in visual reasoning evaluation.

\subsection{Evaluation Framework}

Our evaluation framework is designed to systematically assess the reasoning capabilities of models by decomposing their responses into distinct cognitive stages. To achieve this, we extract structured elements from both the ground truth and model-generated answers, facilitating a detailed comparison.  

\subsubsection{Decomposition of Reasoning Stages}
Based on the ground truth, we categorize the reasoning process into three key stages:

\noindent \textbf{Recognition} (\( R \)): Identifying key visual elements necessary for solving the problem.  
\noindent \textbf{Abstraction} (\( A \)): Inferring higher-level patterns or relationships from \( R \).  
\noindent \textbf{Deduction} (\( D \)): Applying logical reasoning to the abstracted patterns to reach a conclusion.

For model responses, we further introduce a \textbf{Perception} (\( \hat{P} \)) stage, which represents the raw visual elements detected by the model before recognition and reasoning take place. This leads to an extracted model reasoning path:  

\[
\hat{h} = \{ \hat{P}, \hat{R}, \hat{A}, \hat{D} \}
\]

\begin{table*}[!ht]
    \centering
    \renewcommand{\arraystretch}{1.1} 
    \setlength{\tabcolsep}{7pt} 
    \begin{tabular}{l c c c c c c c c c c c}
        \toprule
        \rowcolor{gray!20} \textbf{Model} & \textbf{COT} & \textbf{GT} & \textbf{MA} & \textbf{GC} & \textbf{LP} & \textbf{EO} & \textbf{PR} & \textbf{SA} & \textbf{SC} & \textbf{SP} & \textbf{ALL} \\
        \midrule
        \rowcolor{yellow!20} \multicolumn{12}{c}{\textbf{Open-Source MLLMs}} \\
        Qwen2.5-70b & \texttimes & 0.210 & 0.180 & 0.230 & 0.120 & 0.240 & 0.110 & 0.170 & 0.205 & 0.190 & 0.195 \\
        QVQ-72b & \checkmark & 0.197 & 0.158 & 0.196 & 0.091 & 0.267 & 0.147 & 0.200 & 0.212 & 0.100 & 0.177  \\
        Mulberry-8b & \checkmark & 0.230 & 0.240 & 0.157 & 0.145 & 0.253 & 0.067 & 0.164 & 0.135 & 0.233 & 0.187 \\

        LLaVA-CoT-11b & \checkmark & 0.280 & 0.200 & 0.300 & 0.150 & 0.260 & 0.100 & 0.140 & 0.190 & 0.220 & 0.210 \\

        \midrule
        \rowcolor{red!20} \multicolumn{12}{c}{\textbf{Proprietary MLLMs}} \\
        GPT-4o & \texttimes & 0.262 & 0.178 & 0.353 & 0.091 & 0.200 & 0.093 & 0.164 & 0.250 & 0.200 & 0.192 \\
        Gemini & \checkmark & 0.311 & 0.185 & 0.275 & 0.145 & 0.280 & 0.093 & 0.109 & 0.115 & 0.267 & 0.193 \\
        OpenAI-o1 & \checkmark & 0.279 & 0.233 & 0.275 & 0.164 & 0.293 & 0.120 & 0.145 & 0.192 & 0.200 & 0.215 \\

        \midrule
    \end{tabular}
    \caption{Compact performance comparison of MLLMs with or without CoT support.}
    \label{acc}
    \vspace{-6mm}
\end{table*}

\subsubsection{Evaluation of Reasoning Stages}
For \( R \), \( A \), and \( D \), we perform direct matching against the ground truth using a strong reasoning model, QWQ\cite{qwq-32b-preview}. This ensures a robust comparison while accounting for linguistic variability in model-generated responses.

\subsubsection{Perception Evaluation}
Unlike \( R \), \( A \), and \( D \), the evaluation of perception (\( \hat{P} \)) lacks a well-defined ground truth, as manually verifying each response is labor-intensive. To address this, we propose two unsupervised evaluation methods:

\noindent \textbf{Common Element Agreement.}  
We prompt a large language model (LLM) to extract common visual elements across all model responses. The agreement score for each model is then computed based on the frequency with which it includes these shared elements, offering a measure of consistency across models.

\noindent \textbf{Inter-Model Perception Similarity.}  
To understand how models interpret visual elements relative to one another, we compute pairwise similarity scores between model responses. This results in an \( n \times n \) matrix that quantifies perceptual alignment between models.

Our evaluation framework provides a scalable and structured approach to assess both the structured reasoning process and raw perception abilities of models. This enables a more comprehensive understanding of their cognitive performance across multiple dimensions. More details and examples will be included in the supplementary material.

\section{Experiments and Results}
In this section, we evaluate the visual reasoning abilities of MLLMs with or without CoT.
During the initial phase of our evaluation, we tested a range of open-source and smaller models (13B, 32B) on a randomly selected set of 40 questions. The results revealed that these models struggled, exhibiting both low accuracy and flawed reasoning. Recognizing these limitations, we refined our approach by selecting the most advanced open-source models specifically designed for reasoning: \llava(11B), \mulberry(8B), and \qvq(72B) \cite{qvq-72b-preview}, alongside Qwen2.5-72B, a cutting-edge vision model. 
For proprietary models, we included GPT-4o, as well as two of the most recent leading models: OpenAI-o1 \cite{openai-o1} and Gemini 2.0 Flash Thinking \cite{gemini-flash-thinking}.

To ensure a fair comparison, all models were evaluated under identical conditions with the same default hyperparameters. The evaluation process consisted of three steps.
\begin{enumerate*}
    \item \textit{Generating Responses:} Models produce answers to multi-choice questions.
    \item \textit{Computing Multi-Choice Answer Accuracy:} Accuracy is measured across reasoning categories.
    \item \textit{Assessing Reasoning Path:} Evaluates reasoning based on our proposed evaluation metrics.
\end{enumerate*}

\begin{figure*}[!ht]
    \centering
    \includegraphics[width=\textwidth]{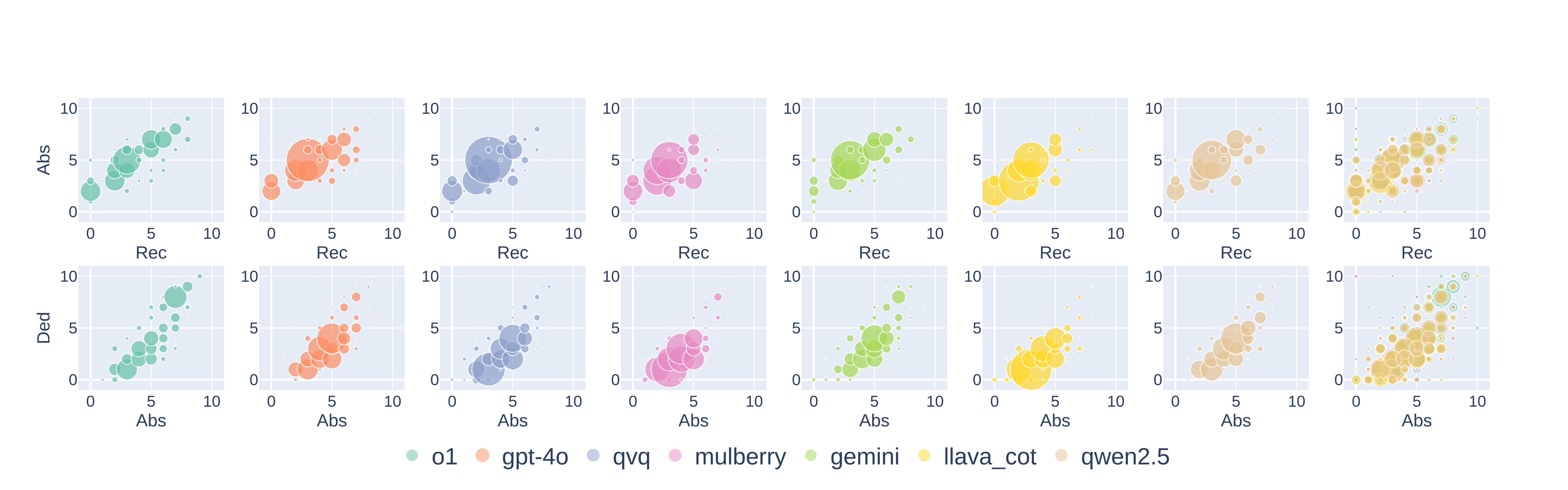}
    \caption{The first row visualizes the correlation between Recognition (\( R \)) and Abstraction (\( A \)), while the second examines the transition from Abstraction (\( A \)) to Deduction (\( D \)). Bubble sizes represent confidence levels. The analysis highlights inter-model variability, with stronger correlations in \( R \rightarrow A \) than \( A \rightarrow D \), indicating challenges in applying logical deduction even when abstraction is well-formed. The last column aggregates all models within the same figure, providing an overall trend.}
    \label{fig:scatter}
    \vspace{-1mm}
\end{figure*}

\subsection{Analysis of Accuracy}

With a random-choice baseline of 0.25 for a 1-out-of-4 selection, all models underperform relative to chance.
Our proposed benchmark reveals a significant challenge in uncovering the subtle intricacies of visual reasoning.
Among open-source MLLMs, \llava\ (11B) achieves the highest overall accuracy, closely followed by \qwen\ (72B). Notably, despite lacking Chain-of-Thought (CoT) support, \qwen\ (72B) remains highly competitive. \mulberry\ (8B) and \qvq\ (72B) trail slightly behind. In the proprietary category, \oone\ demonstrates the best performance, surpassing both \fouro\ (0.192) and Gemini 2.0 Flash Thinking (0.193). Interestingly, despite lacking explicit CoT support, GPT-4o still performs at a comparable level to Gemini. We include a detailed \qvq response in the supplementary material to showcase its ability to switch reasoning paths.

\noindent \textbf{Impact of CoT.} Models equipped with CoT reasoning generally exhibit improved performance in reasoning-intensive tasks. Gemini 2.0 Flash Thinking and \oone\ outperform \fouro, reinforcing the hypothesis that step-by-step reasoning enhances accuracy. A similar trend is observed among open-source models, where \llava\ (11B) leads the group in accuracy. Proprietary models continue to lead in performance over open-source alternatives, benefiting from extensive training data, optimized architectures, and greater computational resources. However, among open-source models, \llava\ (11B) emerges as a strong contender, demonstrating balanced accuracy across various benchmarks. Notably, models incorporating Chain-of-Thought (CoT) reasoning consistently excel in complex problem-solving, underscoring the significance of structured, step-by-step methodologies. Despite this trend, some open-source models, such as \qwen\ (72B), achieve results on par with proprietary counterparts, even without explicit CoT support, suggesting that advancements in model CoT strategy can help bridge the performance gap. 
Our analysis indicates that prioritizing the quality of the reasoning process can reveal crucial diagnostic insights that overall accuracy alone does not capture. In this work, the dataset and benchmark are strategically designed to highlight discrepancies along the reasoning path, thereby offering a robust framework to assess and enhance model performance.

\subsection{Analysis of Reasoning Path}

The result of matching between model over Recognition, Abstraction, and Deduction as shown in Figure \ref{fig:scatter}, providing insight into how well models transition from one stage to the next. The first row compares Recognition to Abstraction, evaluating how effectively models generalize visual elements into higher-level patterns. The second row compares Abstraction to Deduction, assessing the models’ ability to apply logical reasoning to abstracted representations. The bubble size represents the count of the same score.

\noindent \textbf{Recognition vs. Abstraction.}  
The scatter plots in the first row reveal a strong positive correlation between Recognition (R) and Abstraction (A), highlighting that models adept at recognizing key visual elements also tend to excel at forming higher-level abstractions. However, models like GPT-4o and Mulberry exhibit greater variance, suggesting inconsistency in abstraction despite strong recognition capabilities. This inconsistency may indicate limitations in generalizing patterns beyond direct visual cues.  In contrast, models such as \qvq\ and \gemini\ demonstrate more stable performance, implying a robust ability to translate recognized elements into coherent abstract representations. This consistency makes them more reliable in tasks requiring structured pattern recognition. 

\begin{figure}[!h]
    \centering
    \includegraphics[width=0.7\linewidth]{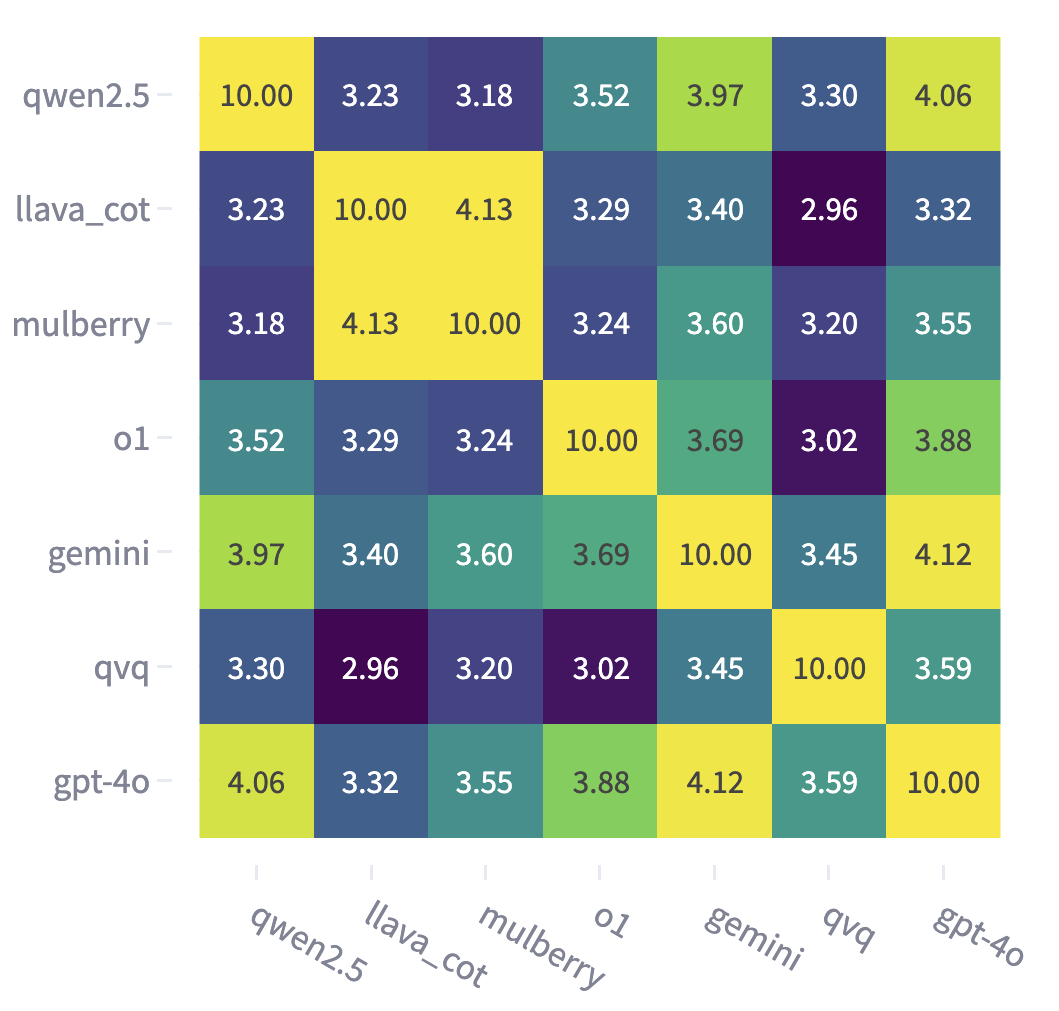}
    \caption{Pairwise Similarity Matrix of Model Responses}
    \label{fig:matrix}
    \vspace{-4mm}
\end{figure}
\begin{figure}[!h]
    \centering
    \includegraphics[width=\linewidth]{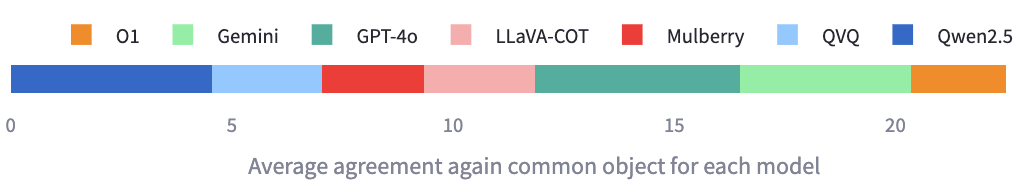}
    \caption{Agreement scores compare each model’s consistency in identifying common visual elements. }
    \label{fig:agree}
    \vspace{-4mm}
\end{figure}
\noindent \textbf{Abstraction vs. Deduction.}  
The second row illustrates the relationship between Abstraction (A) and Deduction (D), revealing a slightly weaker correlation in some models. This suggests that even when models successfully abstract information, their ability to apply logical deduction remains inconsistent.  Notably, \fouro\ and \qvq\ show high variance in abstraction-to-deduction transitions, indicating unpredictability in logical reasoning. This suggests that while these models can form abstract representations, their deductive processes do not always follow a clear or systematic pattern.  Conversely, models like \llava\ and \qwen\ maintain a more compact distribution, signifying a structured and consistent approach to deduction when abstraction is well-formed. Their ability to apply logical reasoning in a predictable manner makes them more reliable for tasks requiring systematic inference.

\noindent \textbf{Perception similarity between models.} As shown in Figure \ref{fig:matrix}, \llava\ and \mulberry\ (4.13), as well as \gemini\ and \fouro\ (4.12), exhibit the strongest perceptual alignment, suggesting these pairs process visual data similarly.
\qvq\ shows the weakest alignment with \llava and \oone, indicating a distinct interpretation of visual inputs. \fouro\ and \qwen\ show relatively strong alignment with multiple models, suggesting a broader generalization. 
\oone\ maintains moderate alignment across models but does not dominate in similarity with any particular one.
Performance trends indicate that proprietary models hold an edge, but open-source models like \llava\ and \qwen\ remain highly competitive.

\noindent \textbf{Perception agreement between models.}  As shown in Figure\ref{fig:agree}, the agreement score provides insight into the consistency of different models in identifying common visual elements across responses. Notably, proprietary models such as \fouro\ and \oone\ exhibit higher consistency, as indicated by their larger proportional representation in the agreement, suggesting that these models more frequently include commonly mentioned elements in the responses. Moreover, models like \qwen and \qvq demonstrate moderate agreement, indicating a balance between consistency and diversity in interpretations.

\section{Conclusion}

In this paper, we introduce \name, the first benchmark designed to assess the fidelity of visual reasoning paths in MLLMs. Through comprehensive experiments and analysis, we highlight the limitations of existing reasoning path evaluation methods and propose new metrics for a more thorough assessment of leading models. Our findings reveal significant shortcomings in current MLLMs’ ability to perform visual reasoning, emphasizing their tendency to excel in either perceptual grounding or logical reasoning, but rarely both. While models with CoT reasoning generally achieve higher accuracy, challenges persist in the transition from abstraction to deduction, limiting overall reasoning capabilities. Additionally, we observe strong correlations between recognition and abstraction, but inconsistencies arise in models’ ability to apply logical inference systematically. By providing a structured evaluation framework, we aim to bridge this gap and pave the way for future advancements in MLLM development.

\newpage
{
    \small
    \bibliographystyle{ieeenat_fullname}
    \bibliography{main}

\begin{thebibliography}{62}
\providecommand{\natexlab}[1]{#1}
\providecommand{\url}[1]{\texttt{#1}}
\expandafter\ifx\csname urlstyle\endcsname\relax
  \providecommand{\doi}[1]{doi: #1}\else
  \providecommand{\doi}{doi: \begingroup \urlstyle{rm}\Url}\fi

\bibitem[gem()]{gemini-flash-thinking}
Gemini 2.0 flash thinking.

\bibitem[ope()]{openai-o1}
Learning to reason with llms.

\bibitem[Agrawal et~al.(2016)Agrawal, Lu, Antol, Mitchell, Zitnick, Batra, and Parikh]{agrawal2016vqavisualquestionanswering}
Aishwarya Agrawal, Jiasen Lu, Stanislaw Antol, Margaret Mitchell, C.~Lawrence Zitnick, Dhruv Batra, and Devi Parikh.
\newblock Vqa: Visual question answering, 2016.

\bibitem[Baradel et~al.(2020)Baradel, Neverova, Mille, Mori, and Wolf]{baradel2020cophycounterfactuallearningphysical}
Fabien Baradel, Natalia Neverova, Julien Mille, Greg Mori, and Christian Wolf.
\newblock Cophy: Counterfactual learning of physical dynamics, 2020.

\bibitem[Barsalou(1999)]{barsalou1999perceptual}
Lawrence~W Barsalou.
\newblock Perceptual symbol systems.
\newblock \emph{Behavioral and brain sciences}, 22\penalty0 (4):\penalty0 577--660, 1999.

\bibitem[Bi et~al.(2024)Bi, Tang, Song, Vosoughi, Nguyen, and Xu]{bi2024eagle}
Jing Bi, Yunlong Tang, Luchuan Song, Ali Vosoughi, Nguyen Nguyen, and Chenliang Xu.
\newblock Eagle: Egocentric aggregated language-video engine.
\newblock \emph{arXiv preprint arXiv:2409.17523}, 2024.

\bibitem[Caffagni et~al.(2024)Caffagni, Cocchi, Barsellotti, Moratelli, Sarto, Baraldi, Cornia, and Cucchiara]{caffagni2024revolution}
Davide Caffagni, Federico Cocchi, Luca Barsellotti, Nicholas Moratelli, Sara Sarto, Lorenzo Baraldi, Marcella Cornia, and Rita Cucchiara.
\newblock The revolution of multimodal large language models: a survey.
\newblock \emph{arXiv preprint arXiv:2402.12451}, 2024.

\bibitem[Cai et~al.(2025)Cai, Yang, and Hu]{cai2025mmiqbenchmarkinghumanlikeabstraction}
Huanqia Cai, Yijun Yang, and Winston Hu.
\newblock Mm-iq: Benchmarking human-like abstraction and reasoning in multimodal models, 2025.

\bibitem[Chollet(2019)]{chollet2019measure}
Fran{\c{c}}ois Chollet.
\newblock On the measure of intelligence.
\newblock \emph{arXiv preprint arXiv:1911.01547}, 2019.

\bibitem[Cobbe et~al.(2021)Cobbe, Kosaraju, Bavarian, Chen, Jun, Kaiser, Plappert, Tworek, Hilton, Nakano, et~al.]{cobbe2021training}
Karl Cobbe, Vineet Kosaraju, Mohammad Bavarian, Mark Chen, Heewoo Jun, Lukasz Kaiser, Matthias Plappert, Jerry Tworek, Jacob Hilton, Reiichiro Nakano, et~al.
\newblock Training verifiers to solve math word problems.
\newblock \emph{arXiv preprint arXiv:2110.14168}, 2021.

\bibitem[Dong et~al.(2024)Dong, Li, Wu, Wang, Zhang, and Guo]{dong2024benchmarkingimprovingimagecaption}
Hongyuan Dong, Jiawen Li, Bohong Wu, Jiacong Wang, Yuan Zhang, and Haoyuan Guo.
\newblock Benchmarking and improving detail image caption, 2024.

\bibitem[Duan et~al.(2021)Duan, Jian, and Tan]{duan2021spacesimulatorphysicalinteractions}
Jiafei Duan, Samson Yu~Bai Jian, and Cheston Tan.
\newblock Space: A simulator for physical interactions and causal learning in 3d environments, 2021.

\bibitem[Fleuret et~al.(2011)Fleuret, Li, Dubout, Wampler, Yantis, and Geman]{fleuret2011comparing}
Fran{\c{c}}ois Fleuret, Ting Li, Charles Dubout, Emma~K Wampler, Steven Yantis, and Donald Geman.
\newblock Comparing machines and humans on a visual categorization test.
\newblock \emph{Proceedings of the National Academy of Sciences}, 108\penalty0 (43):\penalty0 17621--17625, 2011.

\bibitem[Golovneva et~al.(2022)Golovneva, Chen, Poff, Corredor, Zettlemoyer, Fazel-Zarandi, and Celikyilmaz]{golovneva2022roscoe}
Olga Golovneva, Moya Chen, Spencer Poff, Martin Corredor, Luke Zettlemoyer, Maryam Fazel-Zarandi, and Asli Celikyilmaz.
\newblock Roscoe: A suite of metrics for scoring step-by-step reasoning.
\newblock \emph{arXiv preprint arXiv:2212.07919}, 2022.

\bibitem[Goyal et~al.(2017)Goyal, Khot, Summers-Stay, Batra, and Parikh]{goyal2017makingvvqamatter}
Yash Goyal, Tejas Khot, Douglas Summers-Stay, Dhruv Batra, and Devi Parikh.
\newblock Making the v in vqa matter: Elevating the role of image understanding in visual question answering, 2017.

\bibitem[Hendrycks et~al.(2021)Hendrycks, Burns, Kadavath, Arora, Basart, Tang, Song, and Steinhardt]{hendrycks2021measuring}
Dan Hendrycks, Collin Burns, Saurav Kadavath, Akul Arora, Steven Basart, Eric Tang, Dawn Song, and Jacob Steinhardt.
\newblock Measuring mathematical problem solving with the math dataset.
\newblock \emph{arXiv preprint arXiv:2103.03874}, 2021.

\bibitem[Hua et~al.(2024{\natexlab{a}})Hua, Tang, Xu, and Luo]{hua2024v2xum}
Hang Hua, Yunlong Tang, Chenliang Xu, and Jiebo Luo.
\newblock V2xum-llm: Cross-modal video summarization with temporal prompt instruction tuning.
\newblock \emph{arXiv preprint arXiv:2404.12353}, 2024{\natexlab{a}}.

\bibitem[Hua et~al.(2024{\natexlab{b}})Hua, Tang, Zeng, Cao, Yang, He, Xu, and Luo]{hua2024mmcomposition}
Hang Hua, Yunlong Tang, Ziyun Zeng, Liangliang Cao, Zhengyuan Yang, Hangfeng He, Chenliang Xu, and Jiebo Luo.
\newblock Mmcomposition: Revisiting the compositionality of pre-trained vision-language models.
\newblock \emph{arXiv preprint arXiv:2410.09733}, 2024{\natexlab{b}}.

\bibitem[Hudson and Manning(2019)]{hudson2019gqanewdatasetrealworld}
Drew~A. Hudson and Christopher~D. Manning.
\newblock Gqa: A new dataset for real-world visual reasoning and compositional question answering, 2019.

\bibitem[Ittelson(1960)]{ittelson1960visual}
William~H Ittelson.
\newblock Visual space perception.
\newblock 1960.

\bibitem[Janny et~al.(2022)Janny, Baradel, Neverova, Nadri, Mori, and Wolf]{janny2022filteredcophyunsupervisedlearningcounterfactual}
Steeven Janny, Fabien Baradel, Natalia Neverova, Madiha Nadri, Greg Mori, and Christian Wolf.
\newblock Filtered-cophy: Unsupervised learning of counterfactual physics in pixel space, 2022.

\bibitem[Jiang et~al.(2025)Jiang, Zhang, Guo, Li, Qi, Chen, Wang, Jin, Guo, Yan, et~al.]{jiang2025mme}
Dongzhi Jiang, Renrui Zhang, Ziyu Guo, Yanwei Li, Yu Qi, Xinyan Chen, Liuhui Wang, Jianhan Jin, Claire Guo, Shen Yan, et~al.
\newblock Mme-cot: Benchmarking chain-of-thought in large multimodal models for reasoning quality, robustness, and efficiency.
\newblock \emph{arXiv preprint arXiv:2502.09621}, 2025.

\bibitem[Jiang et~al.(2023)Jiang, Ma, Nie, Yu, Zhu, Zhu, and Anandkumar]{jiang2023bongardhoibenchmarkingfewshotvisual}
Huaizu Jiang, Xiaojian Ma, Weili Nie, Zhiding Yu, Yuke Zhu, Song-Chun Zhu, and Anima Anandkumar.
\newblock Bongard-hoi: Benchmarking few-shot visual reasoning for human-object interactions, 2023.

\bibitem[Jiang et~al.(2024)Jiang, Zhang, Sun, Sourati, Ahrabian, Ma, Ilievski, and Pujara]{jiang2024marvelmultidimensionalabstractionreasoning}
Yifan Jiang, Jiarui Zhang, Kexuan Sun, Zhivar Sourati, Kian Ahrabian, Kaixin Ma, Filip Ilievski, and Jay Pujara.
\newblock Marvel: Multidimensional abstraction and reasoning through visual evaluation and learning, 2024.

\bibitem[Johnson et~al.(2017)Johnson, Hariharan, Van Der~Maaten, Fei-Fei, Lawrence~Zitnick, and Girshick]{johnson2017clevr}
Justin Johnson, Bharath Hariharan, Laurens Van Der~Maaten, Li Fei-Fei, C Lawrence~Zitnick, and Ross Girshick.
\newblock Clevr: A diagnostic dataset for compositional language and elementary visual reasoning.
\newblock In \emph{Proceedings of the IEEE conference on computer vision and pattern recognition}, pages 2901--2910, 2017.

\bibitem[Kosslyn(1996)]{kosslyn1996image}
Stephen~M Kosslyn.
\newblock \emph{Image and brain: The resolution of the imagery debate}.
\newblock MIT press, 1996.

\bibitem[Lake et~al.(2017)Lake, Ullman, Tenenbaum, and Gershman]{lake2017building}
Brenden~M Lake, Tomer~D Ullman, Joshua~B Tenenbaum, and Samuel~J Gershman.
\newblock Building machines that learn and think like people.
\newblock \emph{Behavioral and brain sciences}, 40:\penalty0 e253, 2017.

\bibitem[Li et~al.(2023)Li, Li, Savarese, and Hoi]{li2023blip-2}
Junnan Li, Dongxu Li, Silvio Savarese, and Steven Hoi.
\newblock Blip-2: Bootstrapping language-image pre-training with frozen image encoders and large language models.
\newblock In \emph{International conference on machine learning}, pages 19730--19742. PMLR, 2023.

\bibitem[Liang et~al.(2024)Liang, Xu, Hong, Shang, Wang, Fu, and Liu]{liang2024survey}
Zijing Liang, Yanjie Xu, Yifan Hong, Penghui Shang, Qi Wang, Qiang Fu, and Ke Liu.
\newblock A survey of multimodel large language models.
\newblock In \emph{Proceedings of the 3rd International Conference on Computer, Artificial Intelligence and Control Engineering}, pages 405--409, 2024.

\bibitem[Lightman et~al.(2023)Lightman, Kosaraju, Burda, Edwards, Baker, Lee, Leike, Schulman, Sutskever, and Cobbe]{lightman2023let}
Hunter Lightman, Vineet Kosaraju, Yuri Burda, Harrison Edwards, Bowen Baker, Teddy Lee, Jan Leike, John Schulman, Ilya Sutskever, and Karl Cobbe.
\newblock Let's verify step by step.
\newblock In \emph{The Twelfth International Conference on Learning Representations}, 2023.

\bibitem[Lin et~al.(2024)Lin, Wei, An, Gao, Zou, Luo, Huang, Zhang, and Li]{lin2024draw}
Weifeng Lin, Xinyu Wei, Ruichuan An, Peng Gao, Bocheng Zou, Yulin Luo, Siyuan Huang, Shanghang Zhang, and Hongsheng Li.
\newblock Draw-and-understand: Leveraging visual prompts to enable mllms to comprehend what you want.
\newblock \emph{arXiv preprint arXiv:2403.20271}, 2024.

\bibitem[Liu et~al.(2023)Liu, Li, Wu, and Lee]{liu2023llava}
Haotian Liu, Chunyuan Li, Qingyang Wu, and Yong~Jae Lee.
\newblock Visual instruction tuning.
\newblock \emph{Advances in neural information processing systems}, 36:\penalty0 34892--34916, 2023.

\bibitem[Lu et~al.(2024)Lu, Bansal, Xia, Liu, Li, Hajishirzi, Cheng, Chang, Galley, and Gao]{lu2024mathvistaevaluatingmathematicalreasoning}
Pan Lu, Hritik Bansal, Tony Xia, Jiacheng Liu, Chunyuan Li, Hannaneh Hajishirzi, Hao Cheng, Kai-Wei Chang, Michel Galley, and Jianfeng Gao.
\newblock Mathvista: Evaluating mathematical reasoning of foundation models in visual contexts, 2024.

\bibitem[Maaz et~al.(2023)Maaz, Rasheed, Khan, and Khan]{maaz2023video}
Muhammad Maaz, Hanoona Rasheed, Salman Khan, and Fahad~Shahbaz Khan.
\newblock Video-chatgpt: Towards detailed video understanding via large vision and language models.
\newblock \emph{arXiv preprint arXiv:2306.05424}, 2023.

\bibitem[Ma{\l}ki{\'n}ski and Ma{\'n}dziuk(2025)]{malkinski2025deep}
Miko{\l}aj Ma{\l}ki{\'n}ski and Jacek Ma{\'n}dziuk.
\newblock Deep learning methods for abstract visual reasoning: A survey on raven's progressive matrices.
\newblock \emph{ACM Computing Surveys}, 57\penalty0 (7):\penalty0 1--36, 2025.

\bibitem[Marcus(2018)]{marcus2018deep}
Gary Marcus.
\newblock Deep learning: A critical appraisal.
\newblock \emph{arXiv preprint arXiv:1801.00631}, 2018.

\bibitem[Masry et~al.(2022)Masry, Long, Tan, Joty, and Hoque]{masry2022chartqabenchmarkquestionanswering}
Ahmed Masry, Do~Xuan Long, Jia~Qing Tan, Shafiq Joty, and Enamul Hoque.
\newblock Chartqa: A benchmark for question answering about charts with visual and logical reasoning, 2022.

\bibitem[Nie et~al.(2020)Nie, Yu, Mao, Patel, Zhu, and Anandkumar]{nie2020bongard}
Weili Nie, Zhiding Yu, Lei Mao, Ankit~B Patel, Yuke Zhu, and Anima Anandkumar.
\newblock Bongard-logo: A new benchmark for human-level concept learning and reasoning.
\newblock \emph{Advances in Neural Information Processing Systems}, 33:\penalty0 16468--16480, 2020.

\bibitem[Prasad et~al.(2023)Prasad, Saha, Zhou, and Bansal]{prasad2023receval}
Archiki Prasad, Swarnadeep Saha, Xiang Zhou, and Mohit Bansal.
\newblock Receval: Evaluating reasoning chains via correctness and informativeness.
\newblock \emph{arXiv preprint arXiv:2304.10703}, 2023.

\bibitem[Pylyshyn(2003)]{pylyshyn2003seeing}
Zenon~W Pylyshyn.
\newblock \emph{Seeing and visualizing: It's not what you think}.
\newblock MIT press, 2003.

\bibitem[Saparov and He(2022)]{saparov2022language}
Abulhair Saparov and He He.
\newblock Language models are greedy reasoners: A systematic formal analysis of chain-of-thought.
\newblock \emph{arXiv preprint arXiv:2210.01240}, 2022.

\bibitem[Srivastava et~al.(2022)Srivastava, Rastogi, Rao, Shoeb, Abid, Fisch, Brown, Santoro, Gupta, Garriga-Alonso, et~al.]{srivastava2022beyond}
Aarohi Srivastava, Abhinav Rastogi, Abhishek Rao, Abu Awal~Md Shoeb, Abubakar Abid, Adam Fisch, Adam~R Brown, Adam Santoro, Aditya Gupta, Adri{\`a} Garriga-Alonso, et~al.
\newblock Beyond the imitation game: Quantifying and extrapolating the capabilities of language models.
\newblock \emph{arXiv preprint arXiv:2206.04615}, 2022.

\bibitem[Suzgun et~al.(2022)Suzgun, Scales, Sch{\"a}rli, Gehrmann, Tay, Chung, Chowdhery, Le, Chi, Zhou, et~al.]{suzgun2022challenging}
Mirac Suzgun, Nathan Scales, Nathanael Sch{\"a}rli, Sebastian Gehrmann, Yi Tay, Hyung~Won Chung, Aakanksha Chowdhery, Quoc~V Le, Ed~H Chi, Denny Zhou, et~al.
\newblock Challenging big-bench tasks and whether chain-of-thought can solve them.
\newblock \emph{arXiv preprint arXiv:2210.09261}, 2022.

\bibitem[Szegedy et~al.(2013)Szegedy, Zaremba, Sutskever, Bruna, Erhan, Goodfellow, and Fergus]{szegedy2013intriguing}
Christian Szegedy, Wojciech Zaremba, Ilya Sutskever, Joan Bruna, Dumitru Erhan, Ian Goodfellow, and Rob Fergus.
\newblock Intriguing properties of neural networks.
\newblock \emph{arXiv preprint arXiv:1312.6199}, 2013.

\bibitem[Tang et~al.(2023)Tang, Bi, Xu, Song, Liang, Wang, Zhang, An, Lin, Zhu, et~al.]{tang2023video}
Yunlong Tang, Jing Bi, Siting Xu, Luchuan Song, Susan Liang, Teng Wang, Daoan Zhang, Jie An, Jingyang Lin, Rongyi Zhu, et~al.
\newblock Video understanding with large language models: A survey.
\newblock \emph{arXiv preprint arXiv:2312.17432}, 2023.

\bibitem[Tang et~al.(2024{\natexlab{a}})Tang, Guo, Hua, Liang, Feng, Li, Mao, Huang, Bi, Zhang, et~al.]{tang2024vidcomposition}
Yunlong Tang, Junjia Guo, Hang Hua, Susan Liang, Mingqian Feng, Xinyang Li, Rui Mao, Chao Huang, Jing Bi, Zeliang Zhang, et~al.
\newblock Vidcomposition: Can mllms analyze compositions in compiled videos?
\newblock \emph{arXiv preprint arXiv:2411.10979}, 2024{\natexlab{a}}.

\bibitem[Tang et~al.(2024{\natexlab{b}})Tang, Shimada, Bi, Feng, Hua, and Xu]{tang2024empowering}
Yunlong Tang, Daiki Shimada, Jing Bi, Mingqian Feng, Hang Hua, and Chenliang Xu.
\newblock Empowering llms with pseudo-untrimmed videos for audio-visual temporal understanding.
\newblock \emph{arXiv preprint arXiv:2403.16276}, 2024{\natexlab{b}}.

\bibitem[Tang et~al.(2025)Tang, Guo, Liu, Wang, Hua, Zhong, Xiao, Huang, Song, Liang, et~al.]{tang2025generative}
Yunlong Tang, Junjia Guo, Pinxin Liu, Zhiyuan Wang, Hang Hua, Jia-Xing Zhong, Yunzhong Xiao, Chao Huang, Luchuan Song, Susan Liang, et~al.
\newblock Generative ai for cel-animation: A survey.
\newblock \emph{arXiv preprint arXiv:2501.06250}, 2025.

\bibitem[Team(2024{\natexlab{a}})]{qvq-72b-preview}
Qwen Team.
\newblock Qvq: To see the world with wisdom, 2024{\natexlab{a}}.

\bibitem[Team(2024{\natexlab{b}})]{qwq-32b-preview}
Qwen Team.
\newblock Qwq: Reflect deeply on the boundaries of the unknown, 2024{\natexlab{b}}.

\bibitem[Tyen et~al.(2023)Tyen, Mansoor, C{\u{a}}rbune, Chen, and Mak]{tyen2023llms}
Gladys Tyen, Hassan Mansoor, Victor C{\u{a}}rbune, Peter Chen, and Tony Mak.
\newblock Llms cannot find reasoning errors, but can correct them given the error location.
\newblock \emph{arXiv preprint arXiv:2311.08516}, 2023.

\bibitem[Van~der Maas et~al.(2021)Van~der Maas, Snoek, and Stevenson]{van2021much}
Han~LJ Van~der Maas, Lukas Snoek, and Claire~E Stevenson.
\newblock How much intelligence is there in artificial intelligence? a 2020 update.
\newblock \emph{Intelligence}, 87:\penalty0 101548, 2021.

\bibitem[Webb et~al.(2020)Webb, Dulberg, Frankland, Petrov, O’Reilly, and Cohen]{webb2020learning}
Taylor Webb, Zachary Dulberg, Steven Frankland, Alexander Petrov, Randall O’Reilly, and Jonathan Cohen.
\newblock Learning representations that support extrapolation.
\newblock In \emph{International conference on machine learning}, pages 10136--10146. PMLR, 2020.

\bibitem[Wu and Xie(2024)]{wu2024v}
Penghao Wu and Saining Xie.
\newblock V?: Guided visual search as a core mechanism in multimodal llms.
\newblock In \emph{Proceedings of the IEEE/CVF Conference on Computer Vision and Pattern Recognition}, pages 13084--13094, 2024.

\bibitem[Yi et~al.(2019)Yi, Gan, Li, Kohli, Wu, Torralba, and Tenenbaum]{yi2019clevrer}
Kexin Yi, Chuang Gan, Yunzhu Li, Pushmeet Kohli, Jiajun Wu, Antonio Torralba, and Joshua~B Tenenbaum.
\newblock Clevrer: Collision events for video representation and reasoning.
\newblock \emph{arXiv preprint arXiv:1910.01442}, 2019.

\bibitem[Zang et~al.(2024)Zang, Li, Han, Zhou, and Loy]{zang2024contextual}
Yuhang Zang, Wei Li, Jun Han, Kaiyang Zhou, and Chen~Change Loy.
\newblock Contextual object detection with multimodal large language models.
\newblock \emph{International Journal of Computer Vision}, pages 1--19, 2024.

\bibitem[Zeng et~al.(2023)Zeng, Chen, Liu, Jiang, and Jia]{zeng2023mr}
Zhongshen Zeng, Pengguang Chen, Shu Liu, Haiyun Jiang, and Jiaya Jia.
\newblock Mr-gsm8k: A meta-reasoning benchmark for large language model evaluation.
\newblock \emph{arXiv preprint arXiv:2312.17080}, 2023.

\bibitem[Zerroug et~al.(2022)Zerroug, Vaishnav, Colin, Musslick, and Serre]{zerroug2022benchmark}
Aimen Zerroug, Mohit Vaishnav, Julien Colin, Sebastian Musslick, and Thomas Serre.
\newblock A benchmark for compositional visual reasoning.
\newblock \emph{arXiv preprint arXiv:2206.05379}, 2022.

\bibitem[Zhang et~al.(2019)Zhang, Gao, Jia, Zhu, and Zhu]{zhang2019raven}
Chi Zhang, Feng Gao, Baoxiong Jia, Yixin Zhu, and Song-Chun Zhu.
\newblock Raven: A dataset for relational and analogical visual reasoning.
\newblock In \emph{Proceedings of the IEEE/CVF conference on computer vision and pattern recognition}, pages 5317--5327, 2019.

\bibitem[Zhang et~al.(2023)Zhang, Li, and Bing]{zhang2023videollama}
Hang Zhang, Xin Li, and Lidong Bing.
\newblock Video-llama: An instruction-tuned audio-visual language model for video understanding.
\newblock \emph{arXiv preprint arXiv:2306.02858}, 2023.

\bibitem[Zhang et~al.(2024)Zhang, Zhang, Li, Zhao, Karypis, and Smola]{zhang2024multimodalchainofthoughtreasoninglanguage}
Zhuosheng Zhang, Aston Zhang, Mu Li, Hai Zhao, George Karypis, and Alex Smola.
\newblock Multimodal chain-of-thought reasoning in language models, 2024.

\bibitem[Zhu et~al.(2023)Zhu, Chen, Shen, Li, and Elhoseiny]{zhu2023minigpt-4}
Deyao Zhu, Jun Chen, Xiaoqian Shen, Xiang Li, and Mohamed Elhoseiny.
\newblock Minigpt-4: Enhancing vision-language understanding with advanced large language models.
\newblock \emph{arXiv preprint arXiv:2304.10592}, 2023.

\end{thebibliography}
}

\clearpage
\setcounter{page}{1}
\maketitlesupplementary
\setcounter{section}{0} 
\setcounter{figure}{0} 
\section{Qualitative result}

Below is a detailed case study of the question, including reasoning paths from different models and human analysis. We identified a few common mistake patterns:
\begin{enumerate*}[label=\Roman*.]
    \item \textit{Misinterpreting Features:} Errors arise from incorrect counting of elements or misjudging numerical patterns.
    \item \textit{Reliance on Intuition:} Mistakes occur when relying on surface-level visual trends instead of logical transformations.
    \item \textit{Failure to Verify All Options:} Incorrect answers result from selecting seemingly correct choices without checking alternatives.
\end{enumerate*}
\begin{figure}[!h]
    \centering
    \includegraphics[width=\linewidth]{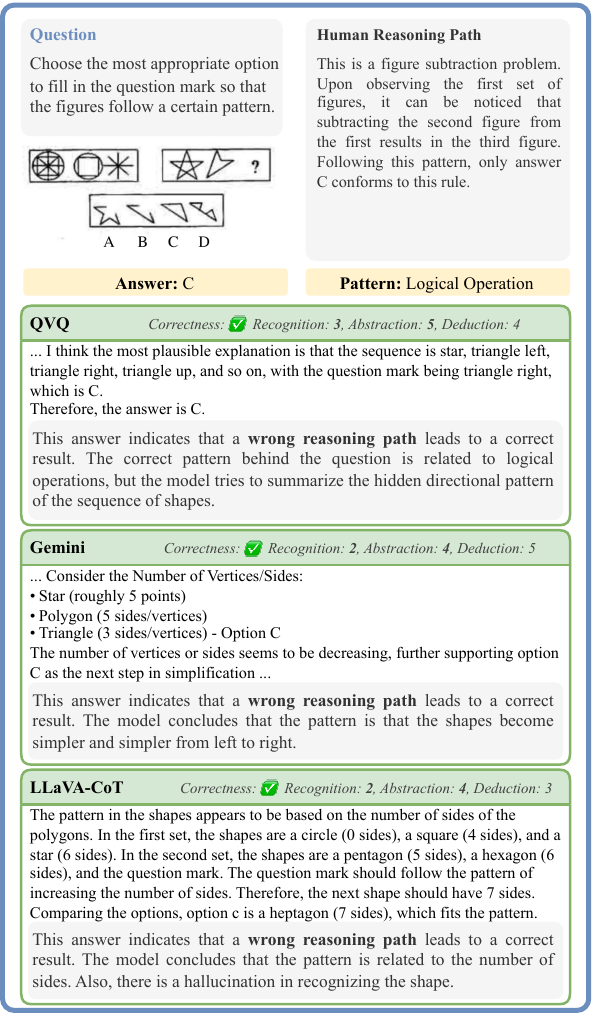}
    \caption{Incorrect reasoning paths still lead to the correct answer}
    \label{fig:enter-label}
\end{figure}
\begin{figure}[!h]
    \centering
    \includegraphics[width=\linewidth]{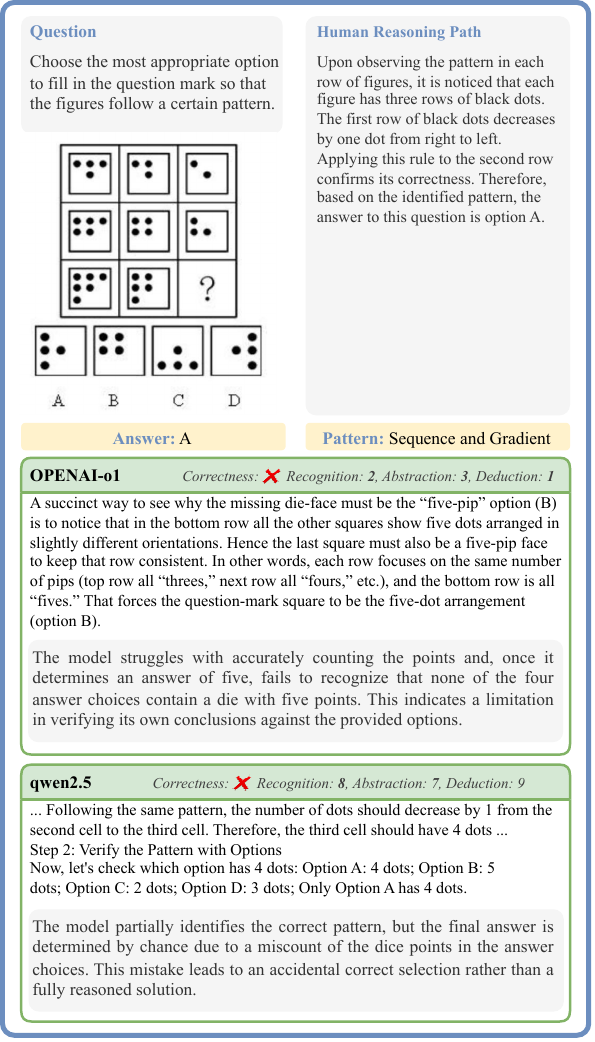}
    \caption{ With human reasoning favoring option A due to a decreasing dot pattern. o1 incorrectly identifies the answer due to miscounting and failing to verify against options, while \qwen accidentally selects the correct answer despite a miscount. }
    \label{fig:enter-label}
\end{figure}

\begin{figure}[!h]
    \centering
    \includegraphics[width=\linewidth]{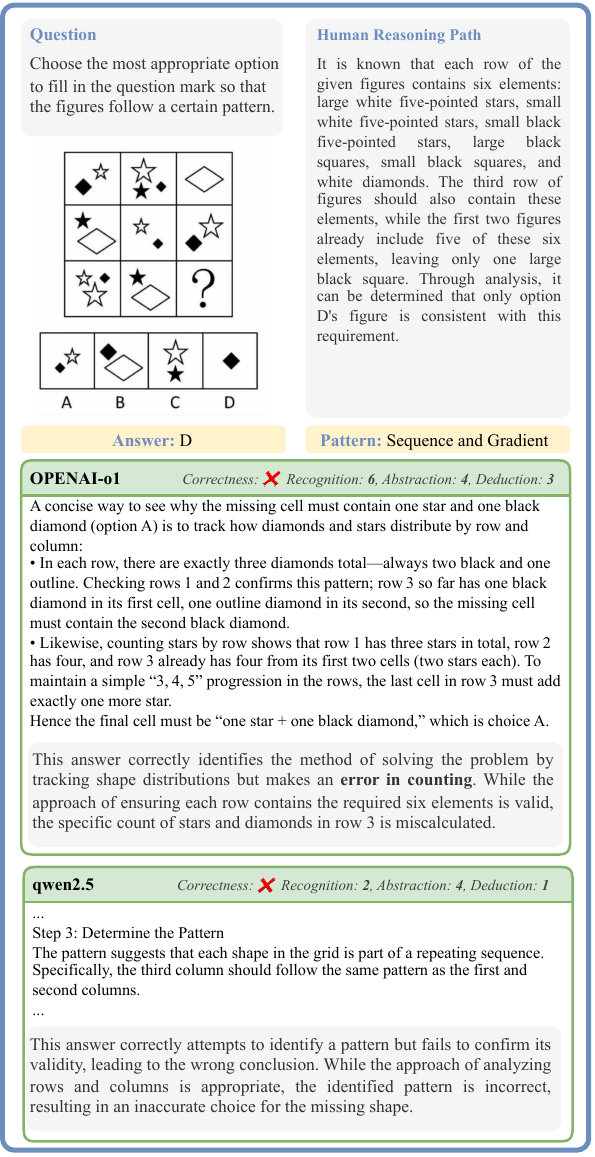}
    \caption{The question asks for the missing figure in a sequence, following a structured pattern of shapes. The correct answer (D) is determined through consistent shape distribution across rows. }
    \label{fig:enter-label}
\end{figure}

\begin{figure}[!h]
    \centering
    \includegraphics[width=\linewidth]{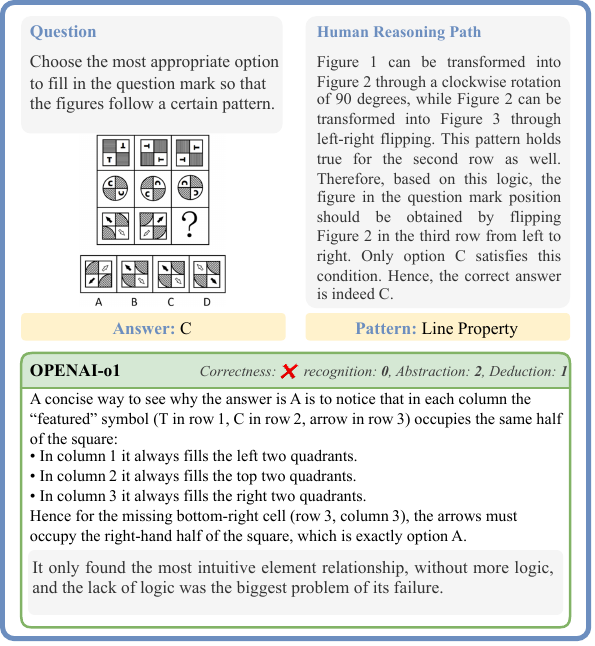}
    \caption{A logical pattern recognition problem where the missing figure follows a rotation and flipping rule. \oone incorrectly chose A by relying on an intuitive pattern without full logical reasoning.}
    \label{fig:enter-label}
\end{figure}

\begin{figure}[!h]
    \centering
    \includegraphics[width=\linewidth]{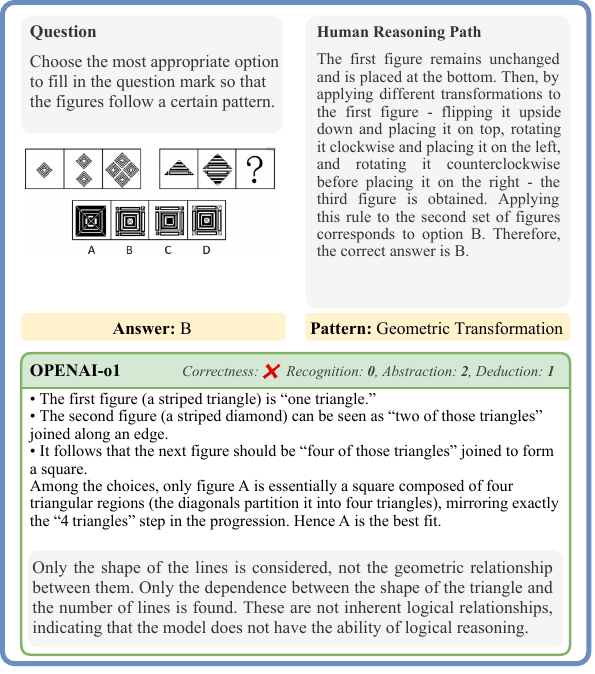}
    \caption{The correct answer is determined through geometric transformations, while \oone incorrectly deduces a numerical pattern in shapes.}
    \label{fig:enter-label}
\end{figure}

\begin{figure}[!h]
    \centering
    \includegraphics[width=\linewidth]{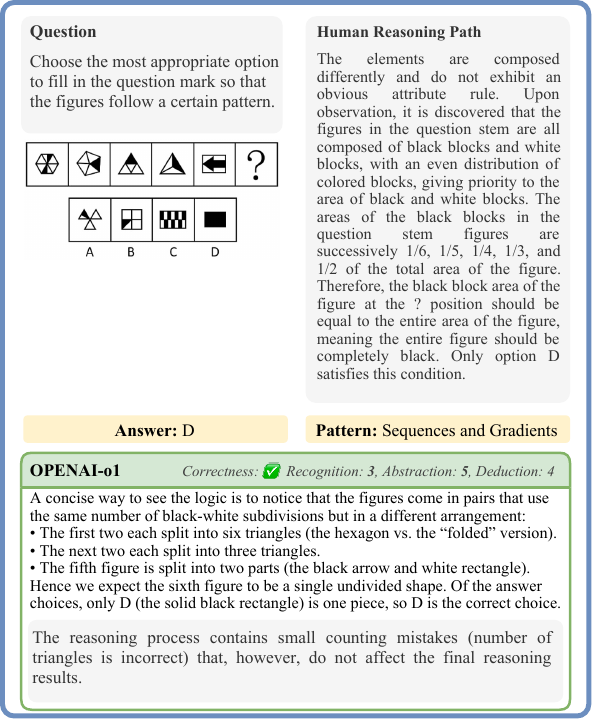}
    \caption{Pattern-based reasoning puzzle where black area increases progressively; the correct answer (D) is a fully black rectangle, completing the sequence.}
    \label{fig:enter-label}
\end{figure}

\begin{figure}[!h]
    \centering
    \includegraphics[width=\linewidth]{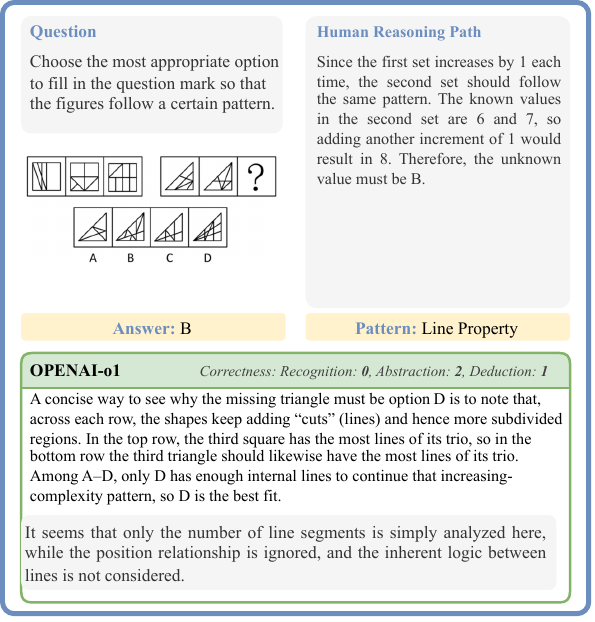}
    \caption{The missing shape follows an incremental line property, with answer choice B completing the sequence logically. }
    \label{fig:enter-label}
\end{figure}

\begin{figure}[!h]
    \centering
    \includegraphics[width=\linewidth]{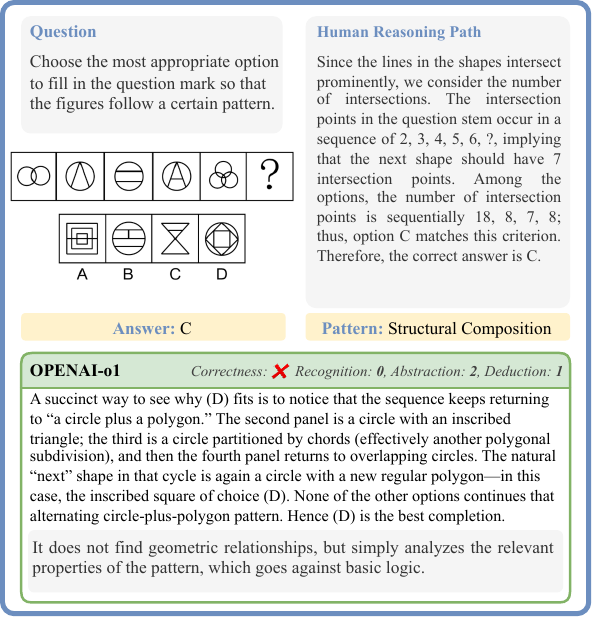}

    \caption{A reasoning-based pattern recognition question where the correct answer (C) follows an intersection count sequence. \oone’s incorrect answer (D) misinterprets the geometric pattern, focusing on shape alternation rather than structural composition.}
    \label{fig:enter-label}
\end{figure}

\begin{figure}[!h]
    \centering
    \includegraphics[width=\linewidth]{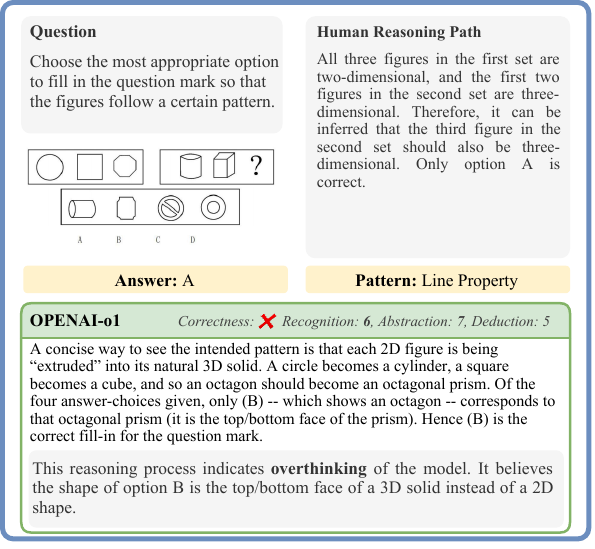}
    \caption{Illustration of a pattern recognition puzzle where 2D shapes transform into their corresponding 3D forms, with the correct answer (A) requiring a three-dimensional figure. \oone overthinks, selecting (B) based on a misinterpretation of shape extrusion.}
    \label{fig:enter-label}
\end{figure}

\begin{figure}[!h]
    \centering
    \includegraphics[width=\linewidth]{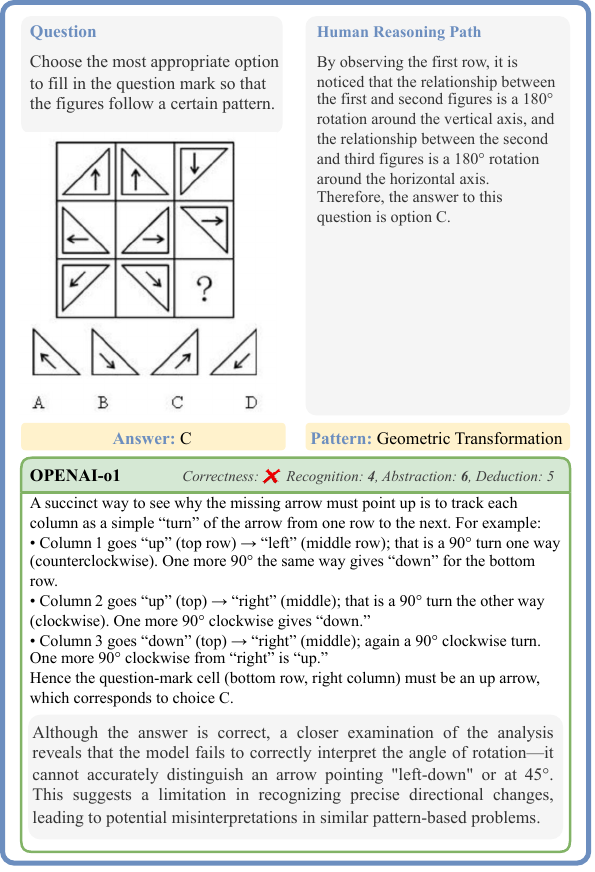}
    \caption{The correct answer (C) follows a sequence of 180° and 90° rotations, though analysis reveals challenges in accurately interpreting directional changes.}
    \label{fig:enter-label}
\end{figure}
\begin{figure}[!h]
    \centering
    \includegraphics[width=1.1\linewidth]{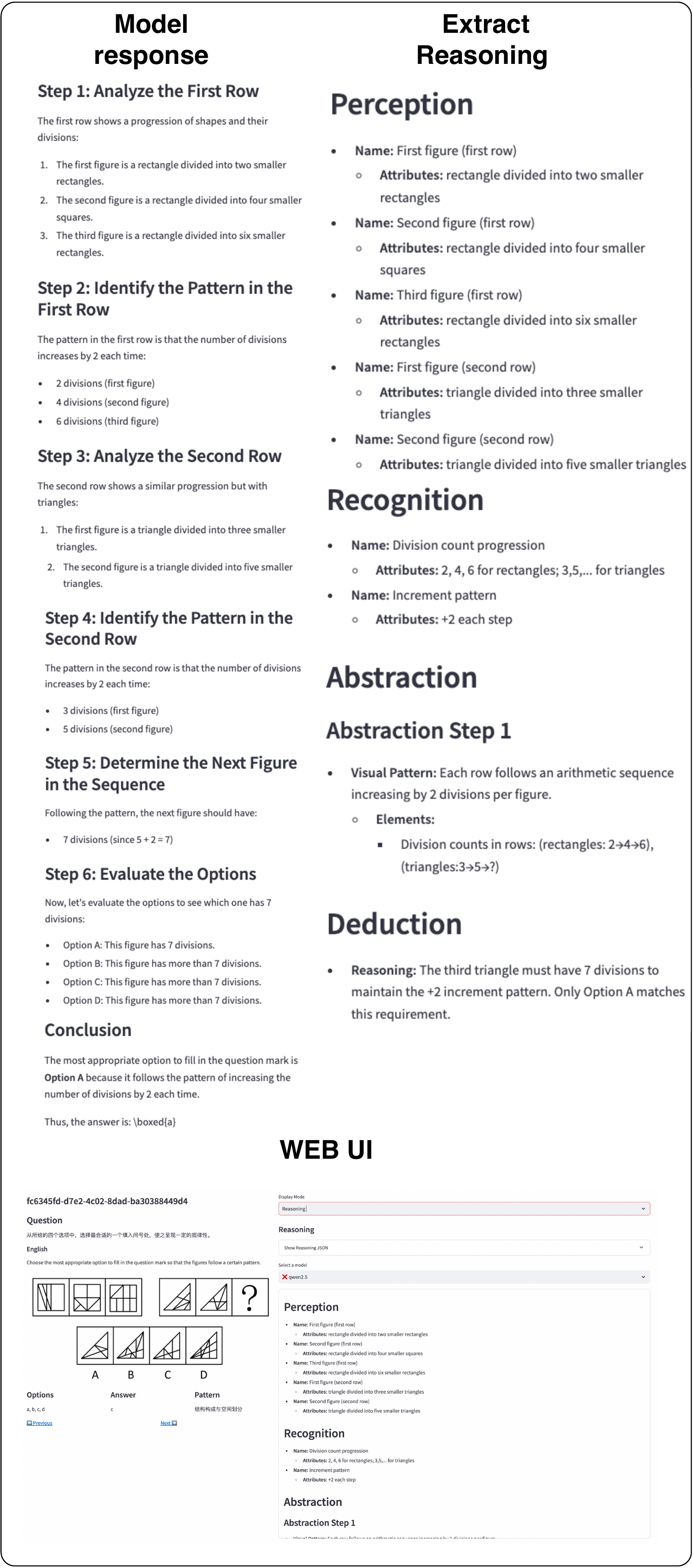}
    \caption{ Example of extracted reasoning and collaborative annotation interface.}
    \label{fig:extexp}
\end{figure}

\clearpage

\begin{table*}[t]
    \centering
    \renewcommand{\arraystretch}{1.1} 
    \setlength{\tabcolsep}{7pt} 
    \begin{tabular}{l c c c c c c c c c c c}
        \toprule
        \rowcolor{gray!20} \textbf{Model} & \textbf{COT} & \textbf{GT} & \textbf{MA} & \textbf{GC} & \textbf{LP} & \textbf{EO} & \textbf{PR} & \textbf{SA} & \textbf{SC} & \textbf{SP} & \textbf{ALL} \\
        \midrule
\rowcolor{yellow!20} \multicolumn{12}{c}{\textbf{Open-Source LMMs}} \\
qwen2.5 & \texttimes & 0.230 & 0.179 & 0.235 & 0.091 & 0.240 & 0.080 & 0.182 & 0.115 & 0.241 & 0.174 \\
qvq & \checkmark & 0.226 & 0.172 & 0.208 & 0.104 & 0.290 & 0.172 & 0.234 & 0.268 & 0.107 & 0.199 \\
mulberry & \checkmark & 0.233 & 0.245 & 0.157 & 0.154 & 0.257 & 0.069 & 0.173 & 0.137 & 0.241 & 0.192 \\
llava-cot & \checkmark & 0.180 & 0.186 & 0.235 & 0.091 & 0.253 & 0.160 & 0.093 & 0.216 & 0.267 & 0.184 \\
\midrule
\rowcolor{red!20} \multicolumn{12}{c}{\textbf{Proprietary LMMs}} \\
o1 & \checkmark & 0.279 & 0.236 & 0.280 & 0.164 & 0.293 & 0.120 & 0.145 & 0.196 & 0.207 & 0.217 \\
gemini & \checkmark & 0.311 & 0.189 & 0.275 & 0.148 & 0.280 & 0.093 & 0.109 & 0.115 & 0.267 & 0.195 \\
gpt-4o & \texttimes & 0.262 & 0.178 & 0.360 & 0.096 & 0.200 & 0.093 & 0.167 & 0.255 & 0.207 & 0.194 \\
        \midrule
    \end{tabular}
    \caption{Unlike the previous evaluation, which considered all invalid cases as incorrect, this updated assessment excludes invalid cases from the computation. By removing invalid responses from the accuracy calculation, the reported scores better reflect the models’ true performance on valid instances.}
    \label{newcc}
\end{table*}

\section{Model Comparison}

\begin{figure}[!h]
    \centering
    \includegraphics[width=1\linewidth]{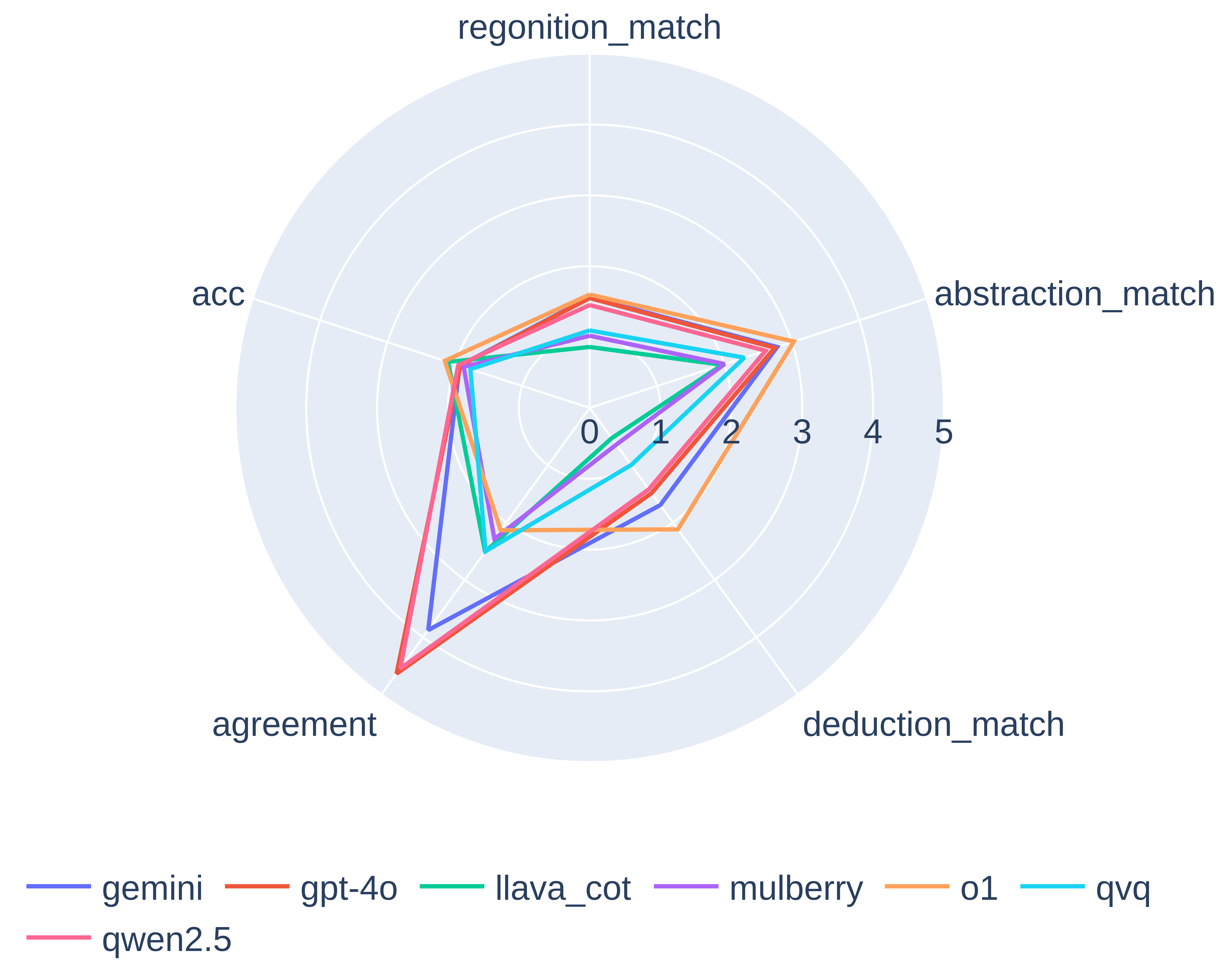}
    \caption{Holistic comparison of Model Reasoning and Perception Abilities}
    \label{fig:radar}
\end{figure}

We also include a radar chart for a quick comparison of model performance, as shown in Figure \ref{fig:radar}. The plot highlights variations across key reasoning and perception dimensions. Notably, models such as Qwen2.5 and GPT-4o exhibit higher agreement scores, suggesting stronger consensus in visual perception. Meanwhile, o1 demonstrates relatively higher abstraction and deduction scores, indicating superior reasoning capabilities. Differences in recognition performance suggest that models vary in their ability to identify key visual elements, which may impact subsequent reasoning stages.

We also present a new data accuracy evaluation in Table \ref{newcc}, which excludes invalid responses and computes the mean of the remaining data. This provides a more accurate assessment of models like QVQ, which utilizes all tokens for reasoning without explicitly producing a final answer.

\section{Data Process}

Our selection process ensures the highest quality and clarity by following strict criteria. First, we remove all ambiguous questions, excluding those with multiple correct answers or requiring external knowledge beyond the provided visual content. To maintain relevance, we only include data from the past 5-8 years, up to 2024. Additionally, we collect problems from various provinces, prioritizing those known for higher difficulty levels, such as Beijing and Shanghai, to ensure a rigorous dataset. Each question is handcrafted and filtered by annotators with at least a master’s degree or higher, guaranteeing clear and precise final selections.

To ensure a clear and structured reasoning path for each problem, we follow a systematic approach. First, annotators attempt to solve the problem independently. If successful, we collect a concise description of their solution process. If an annotator is unable to solve the problem, we provide the correct answer as a hint to determine whether a reasoning path can be constructed. If both steps fail, we assess whether the problem is inherently difficult or if it follows an exceptionally rare pattern. Any problems deemed too rare are removed, and only solutions provided by annotators serve as the gold standard for reasoning clarity.

\section{Example of Data Extraction}
Below, we present the model’s response alongside the extracted reasoning elements, as illustrated in Figure \ref{fig:extexp}. The extracted reasoning captures the core logical structure of the response, ensuring key insights are conveyed effectively. Additionally, we showcase the web-based UI used for collaboration with annotators. We will share these resources with the community to facilitate further research.

\section{Analysis of QVQ}
\subsection{How the reasoning works with visual}
\begin{figure*}[htbp]
    \centering
    \begin{minipage}{0.49\textwidth}
        \centering
        \includegraphics[width=\linewidth]{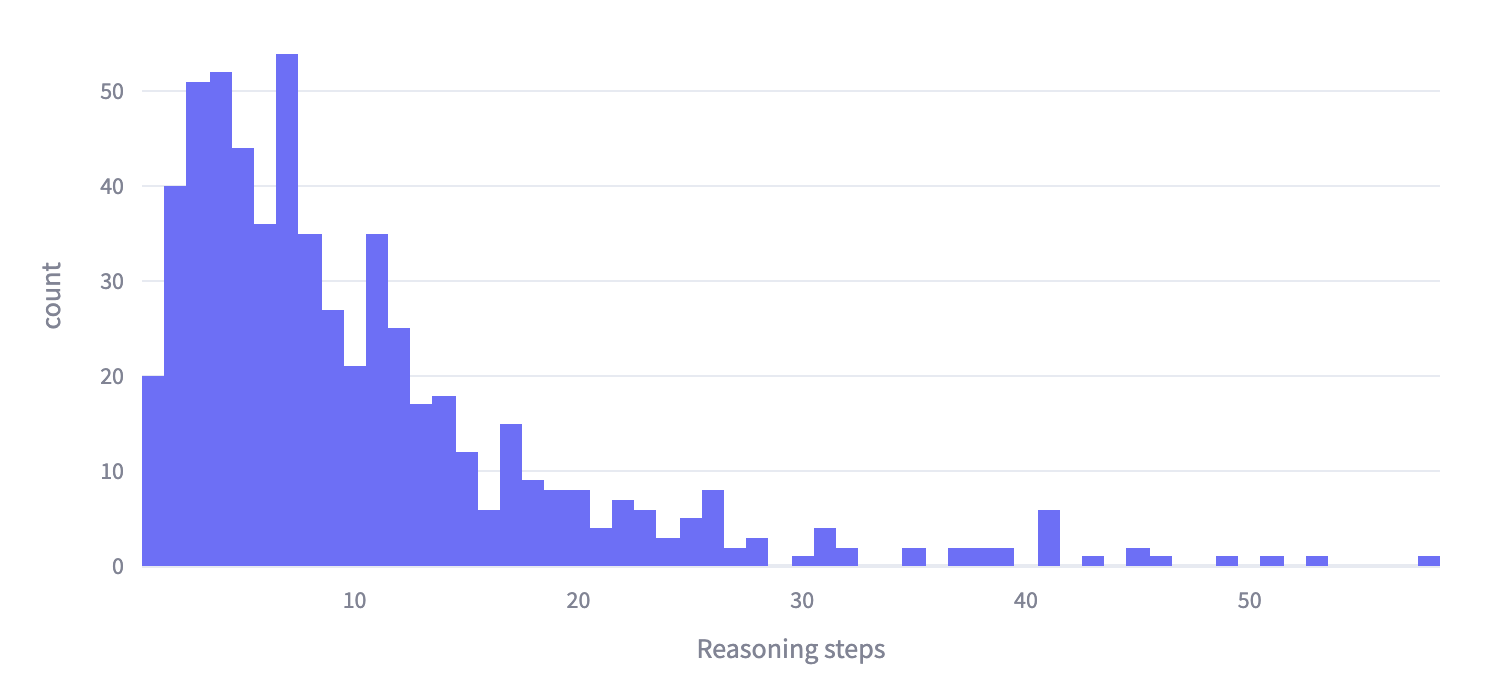}
        \caption{Histogram showing the distribution of reasoning steps. The x-axis represents the number of reasoning steps, while the y-axis represents the count of occurrences.}
        \label{fig:reasoning-steps-count}
    \end{minipage}
    \hfill
    \begin{minipage}{0.49\textwidth}
        \centering
        \includegraphics[width=\linewidth]{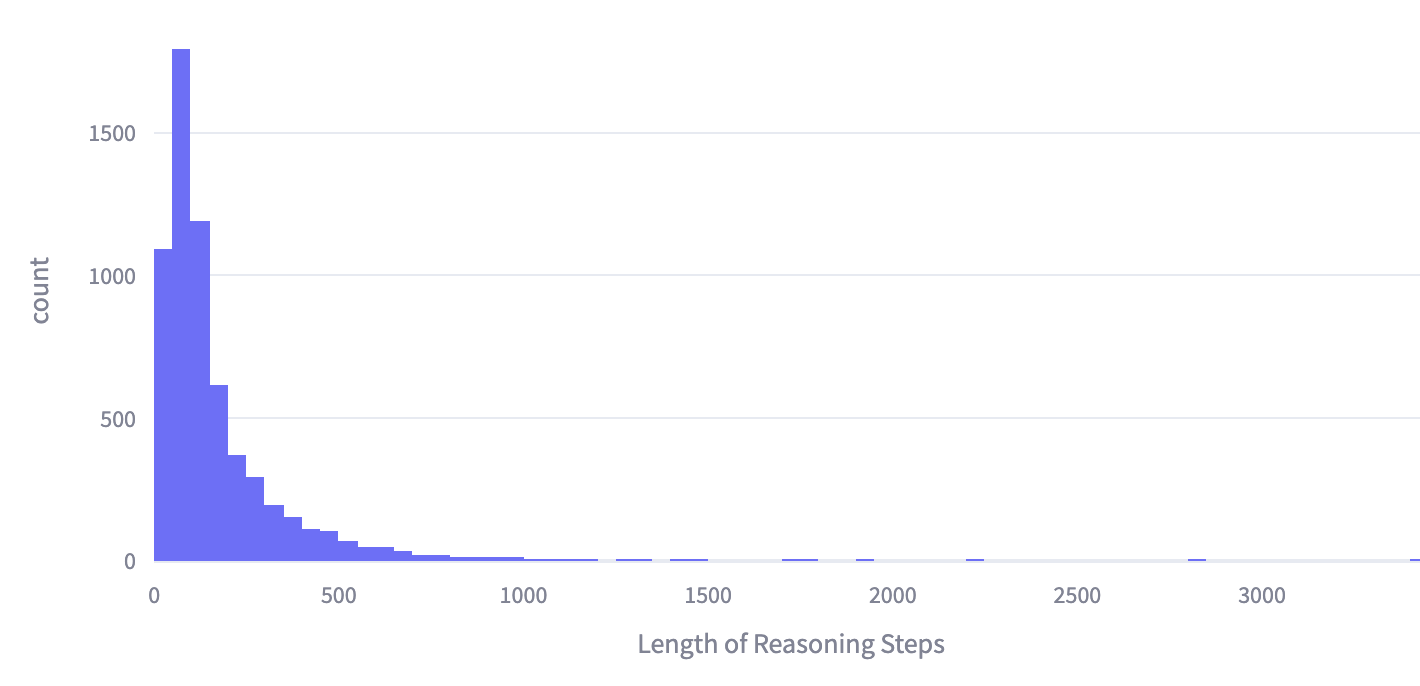}
        \caption{Histogram showing the distribution of the length of reasoning steps in words. The x-axis represents the length of reasoning steps, while the y-axis represents the count of occurrences.}
        \label{fig:reasoning-steps-length}
    \end{minipage}
    \caption{Comparison of reasoning step distributions: (a) Number of steps and (b) Length of steps in words.}
\end{figure*}

\begin{figure*}[!h]
    \centering
    \includegraphics[width=\linewidth]{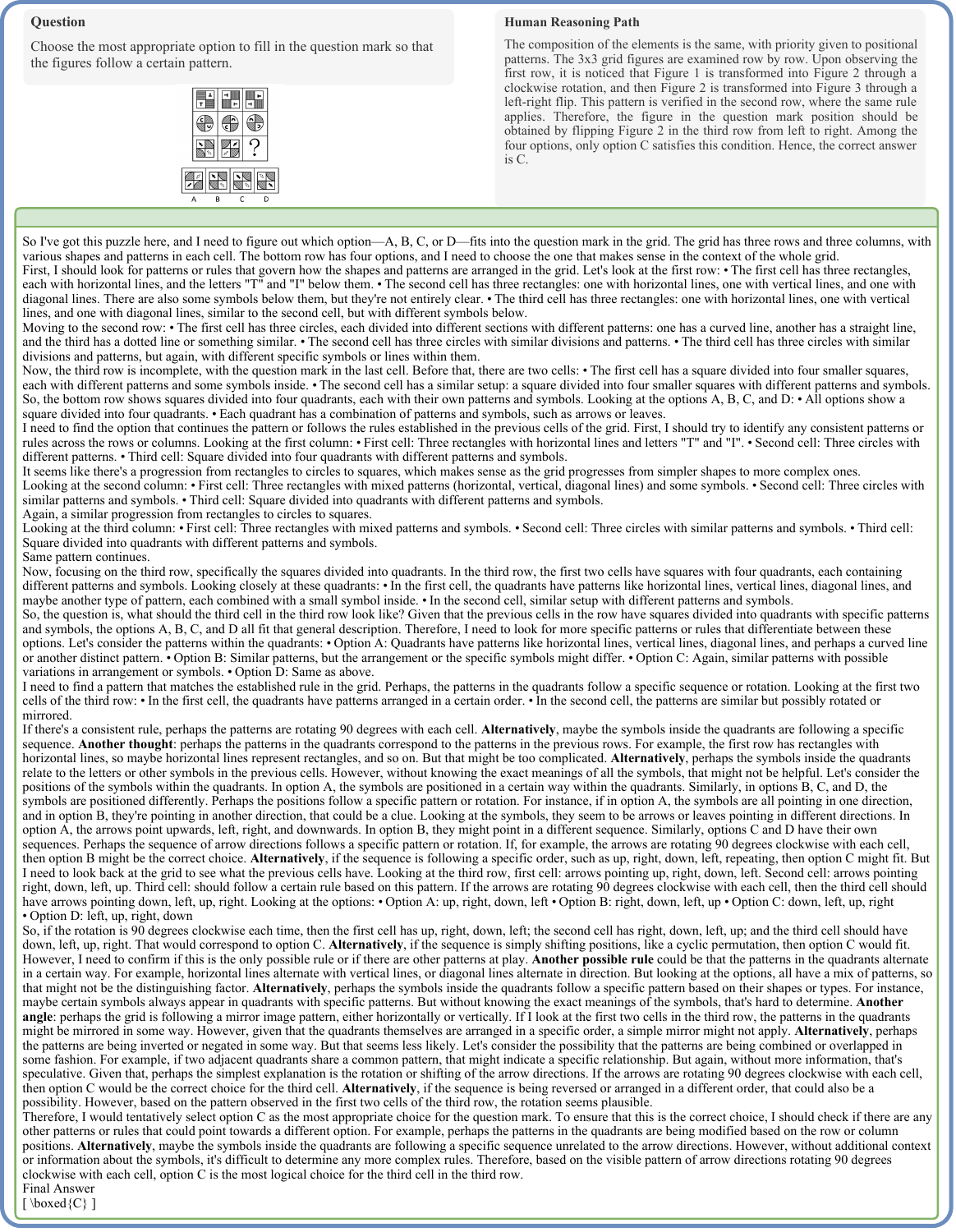}
    \caption{The example reasoning path of QVQ shows that QVQ needs extensive tokens to solve the problem.}
    \label{fig:qvq-example}
\end{figure*}

The reasoning process in multi-modal models remains an open challenge, particularly in proprietary MLLMs such as \oone, which do not reveal their internal reasoning paths. Meanwhile, weaker open-source models lack the ability to explore different strategies when analyzing visual patterns. In contrast, QVQ exhibits a unique characteristic: it systematically attempts multiple reasoning strategies to solve a given problem rather than following a single fixed approach, as shown in Figure \ref{fig:qvq-example}.

Our observations indicate that QVQ genuinely explores various paths to reasoning, often exceeding 10 steps in an effort to find the correct answer.

This characteristic differentiates it from other models, as it does not settle on a single reasoning trajectory prematurely. However, this exhaustive approach contributes to its low performance, as the model frequently exceeds the available 16k context length before converging on a final answer. Instead of providing a clear and concise response, QVQ continues its reasoning process until it runs out of tokens, leading to incomplete or inconclusive outputs.

One notable observation is that QVQ consistently uses “Alternatively” to introduce a shift in reasoning, allowing us to identify alternative considerations as distinct reasoning steps. This enables us to analyze the step count and token distribution for each step.

Building on this observation, Figure~\ref{fig:reasoning-steps-count} reveals that while the majority of QVQ tasks require a relatively small number of reasoning steps, a long-tail distribution is evident, with some tasks demanding an excessive number of steps. This suggests that for more complex problems, QVQ struggles to streamline its reasoning and instead resorts to prolonged deliberation without efficiently converging on a solution. The presence of tasks requiring over 30 or even 50 steps highlights the difficulty of the problem, as the model fails to resolve them within a reasonable step count.

Similarly, Figure~\ref{fig:reasoning-steps-length} illustrates the distribution of token usage per reasoning step. While most steps remain relatively short, the steep long-tail distribution shows that some reasoning steps extend to several hundred or even thousands of tokens. This suggests that QVQ, when faced with difficult problems, often generates disproportionately lengthy reasoning sequences in an attempt to work through them. However, rather than leading to a clear resolution, these extended steps further contribute to inefficiency and excessive token consumption.

Together, these findings indicate that QVQ struggles with complex reasoning tasks, as evidenced by both excessive step counts and highly variable step lengths. The model’s inability to converge efficiently suggests that certain problems remain unsolved within a reasonable context window. This inefficiency highlights a fundamental limitation in QVQ’s reasoning approach—rather than refining its logic, it frequently resorts to verbosity and fragmented steps, ultimately failing to reach a satisfactory conclusion.

Interestingly, while \oone does not return intermediate reasoning steps, we can infer that it undergoes a similar iterative process. Based on token usage, we estimate that O1 requires an average of 6-8k tokens before reaching a decision. This suggests that proprietary models may employ comparable multi-step reasoning strategies but optimize their token allocation more efficiently than QVQ.

Overall, while QVQ demonstrates an impressive ability to explore diverse reasoning strategies, its lack of an effective stopping mechanism and excessive token consumption hinder its performance. Future improvements should focus on optimizing its reasoning efficiency—potentially by refining the model's ability to prioritize promising reasoning paths early on and reducing unnecessary exploration.

\end{document}